\documentclass{article}
\usepackage[T1]{fontenc}
\usepackage{iclr2026_conference,times}
\iclrfinalcopy

\usepackage{amsmath,amsfonts,bm}

\def\eqref#1{equation~\ref{#1}}

\def\1{\bm{1}}

\DeclareMathAlphabet{\mathsfit}{\encodingdefault}{\sfdefault}{m}{sl}
\SetMathAlphabet{\mathsfit}{bold}{\encodingdefault}{\sfdefault}{bx}{n}

\usepackage{hyperref}
\usepackage{url}
\usepackage{array,booktabs}
\usepackage{graphicx}
\usepackage{grffile}
\usepackage{caption}
\usepackage{float}
\usepackage{placeins}
\newcommand{\headerlogheight}{0.38in}
\newcommand{\headerlogheightleft}{0.36in}
\title{A Generative Foundation Model for Multimodal Tokamak Tasks}
\author{%
  Yingshuai Ji$^{1}$ \& Haodong Feng$^{1,*}$ \& Guohong Deng$^{2}$ \& Gelin Bian$^{1}$ \& Yue Wang$^{1,*}$ \\[0.75ex]
  $^{1}$Zhongguancun Academy 
  $^{2}$XinZhuAI \\
  $^{*}$Corresponding authors: \texttt{fenghaodong@bza.edu.cn}, \texttt{yuewang@bza.edu.cn}.
}

\begin{document}

\maketitle
\fancyhead[L]{\includegraphics[height=\headerlogheightleft]{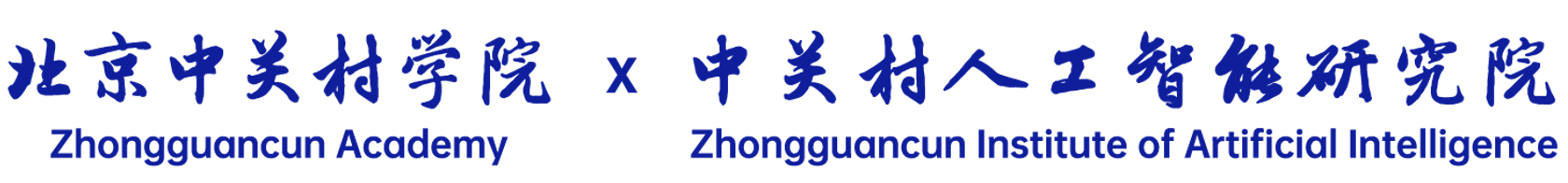}}
\fancyhead[R]{\includegraphics[height=\headerlogheight]{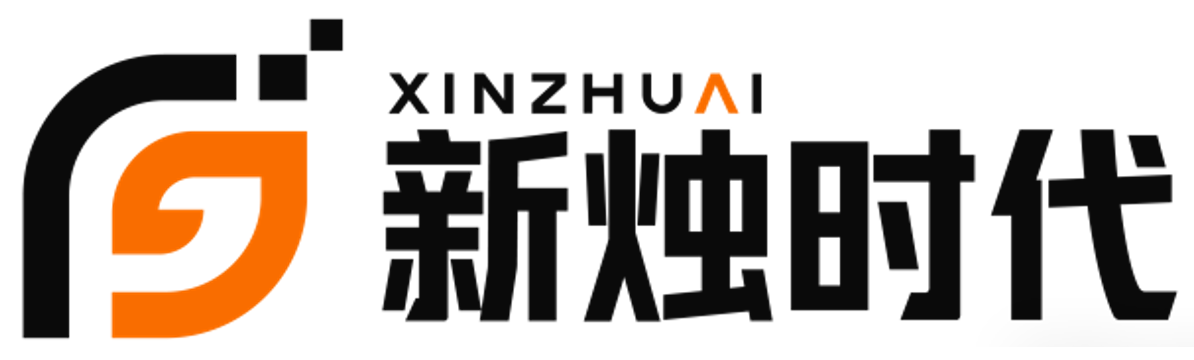}}
\fancyfoot[C]{\thepage}
\renewcommand{\headrulewidth}{0.4pt}
\renewcommand{\footrulewidth}{0pt}
\pagestyle{fancy}
\setlength{\headwidth}{\textwidth}
\thispagestyle{fancy}

\begin{abstract}
Tokamak experiments record diverse measurements, including waveforms, profiles, fields, and scalars, which are used to reconstruct the plasma, forecast its evolution, recover associated control signals, and estimate disruption risk. Most models are built separately for one task and a fixed set of inputs and outputs, and each relies on a prescribed subset of diagnostics rather than structuring the full multimodal observation set to extract global, cross-signal information about the discharge. These limitations motivate a reusable foundation model that adapts to each task's input--output schema and models heterogeneous target signals jointly, rather than retraining a separate model for every configuration. In this work, we extend the TokaMark benchmark with two groups of tasks for actuator action inversion and disruption prediction, yielding 20 MAST tasks in six groups with explicit schemas and evaluation protocols. Then, we introduce GenToka, a generative tokamak foundation model that jointly generates schema-selected targets through a shared conditional diffusion Transformer, with signal-specific codecs preserving native time series, profiles, fields, and scalars; foundation pretraining and task-specific adaptation transfer cross-signal structure across all six groups. Evaluation on MAST measurements shows that GenToka reduces macro-average normalized root mean squared error relative to TokaMind by 4.7\% on Groups~1--4 and by 17.7\% on Group~5, and improves macro-average window-level F1 on Group~6 by 8.4\% at a common validation-selected threshold. GenToka provides a reusable, generative, and multimodal pathway for experimental tokamaks, linking plasma-state reconstruction, actuator inference, and disruption warning from shared measurements to support more integrated, data-driven fusion plasma prediction and control research.
\end{abstract}

\section{Introduction}
\label{sec:introduction}

Controlled nuclear fusion offers a potential source of sustainable energy, and magnetic confinement in tokamaks is a promising route toward this goal \citep{degrave2022control}. Reliable models of tokamak dynamics are needed to reconstruct plasma states, anticipate their evolution, and inform the control of stability, confinement, and magnetic configuration \citep{walker2020tokamak}. These modeling tasks are challenging because the plasma is only partially observed: diagnostic measurements differ in physical meaning, sampling rate, spatial structure, and availability, while plasma evolution also depends on actuator histories and operating conditions. Resolving the coupled processes with physics-based simulations is high computational load, motivating surrogate models that connect these heterogeneous observations to the quantities needed for a particular operational task \citep{humphreys2020fusionml,rousseau2026tokamark}.

Learning tokamak dynamics from experimental measurements provides a complementary way to model these relationships. Data-driven methods have shown strong results in equilibrium reconstruction, kinetic-profile estimation, disruption prediction, and magnetic control \citep{joung2023gsdeepnet,shousha2024diii,katesharbeck2019disruption,degrave2022control}. However, many systems are developed for a fixed signal set, output structure, and prediction horizon, which limits reuse as diagnostic availability and task requirements change. FAIR-MAST and TokaMark have improved access to experimental data and standardized evaluation across forward modeling tasks \citep{jackson2024fairmast,jackson2025opendata,rousseau2026tokamark}, while TokaMind demonstrates the value of reusable multimodal representations \citep{boschi2026tokamind}. Two needs remain: a broader benchmark that also covers disruption risk and inverse actuator inversion, and modeling that can characterize joint uncertainty over heterogeneous diagnostics. Supervised learning is well suited to estimating a conditional mean under a fixed input--output contract, but it returns one point prediction and often factorizes each output channel into a separate regression. Conditional generative models are more expressive in this setting, as they can represent the model dependencies jointly by different tasks. Recent work on fluid systems shows that one conditional diffusion model can span diverse tasks within a physical domain \citep{feng2025fluidzero}, suggesting a parallel generative perspective for heterogeneous tokamak objectives.

In this work, we introduce GenToka, a \textbf{Gen}erative \textbf{Toka}mak Foundation Model for heterogeneous plasma measurements. Each task is specified by a \emph{schema}: the input and output signals, their physical roles, and their temporal support. GenToka encodes the available diagnostic and actuator measurements as context tokens and represents each complete requested output---a time-series window, profile, field, or scalar---by one noisy target token. This gives heterogeneous outputs a common interface for joint generation, while signal-specific heads restore their native structures after shared conditional diffusion Transformer denoising \citep{peebles2023dit,ma2025latte}.

GenToka is trained in two stages. Foundation pretraining learns cross-signal relationships through co-temporal reconstruction with randomly masked context, exposing the model to incomplete observations. Downstream fine-tuning then adapts the pretrained backbone and task-specific interfaces to the required inputs and outputs. We evaluate the predictive accuracy of individual generated outputs on MAST measurements across twenty tasks divided into six groups: equilibrium reconstruction (Group~1), magnetic dynamics (Group~2), profile and diagnostic dynamics (Group~3), long-horizon magnetohydrodynamic (MHD) dynamics (Group~4), actuator action inversion (Group~5), and disruption prediction (Group~6). Their schemas and evaluation protocols are given in Section~\ref{sec:benchmark} and Section~\ref{sec:benchmark-protocol}.

Our contributions are:

\begin{enumerate}
    \item \textbf{Comprehensive benchmark for tokamak tasks.} We define 20 tasks in six groups that connect prediction of plasma responses, actuator action inversion, and disruption prediction from historical measurements. Explicit input--output and temporal definitions enable reproducible evaluation of these complementary capabilities, providing a common basis for assessing the breadth of tokamak foundation models.

    \item \textbf{Generative tokamak foundation model.} GenToka uses a shared diffusion Transformer to model relationships among heterogeneous output signals conditioned on the available measurements. Signal-specific interfaces preserve the native structures of time series, profiles, fields, and scalars while making them accessible to a common generative backbone. This design supports different input--output configurations through one conditional generation framework, allowing cross-signal representations learned during foundation pretraining to be reused across all six task groups.

    \item \textbf{Evaluation on 20 tasks using MAST measurements.} Compared with the existing TokaMind model, GenToka reduces macro-average NRMSE by 4.7\% across Groups 1--4 and by 17.7\% across Group~5. Across Group~6, macro-average window-level F1 improves by 8.4\% at a common validation-selected threshold. Paired comparisons with the same GenToka architecture initialized from scratch show additional gains from foundation pretraining. Together, these results demonstrate both competitive accuracy relative to TokaMind and the benefit of pretraining over scratch initialization across all six task groups.
\end{enumerate}

\section{Related Work}
\label{sec:related-work}

\subsection{Tokamak modeling and scientific foundation models}

Data-driven tokamak research has addressed equilibrium reconstruction, kinetic-profile estimation, disruption prediction, and magnetic control \citep{joung2023gsdeepnet,shousha2024diii,katesharbeck2019disruption,degrave2022control}, with recent work further exploring multimodal reconstruction and transfer across operating settings \citep{jalalvand2025superresolution,zheng2023disruptiontransfer}. These studies demonstrate the value of experimental diagnostics and actuator histories, but most are organized around a prescribed signal set, output structure, and modeling task. FAIR-MAST improves access to structured experimental data \citep{jackson2024fairmast,jackson2025opendata}, while TokaMark standardizes data processing and evaluation across a suite of forward tasks \citep{rousseau2026tokamark}. These developments provide the starting point for our twenty-task benchmark, which spans plasma-response modeling, actuator action inversion, and disruption prediction with explicit input and evaluation protocols.

Foundation models offer a complementary perspective by emphasizing reusable representations and adaptation across tasks and data configurations. Scientific foundation models have pursued such reuse in complementary settings: ClimaX pretrains across heterogeneous climate variables and spatiotemporal coverage \citep{nguyen2023climax}, Poseidon transfers representations across PDE families \citep{herde2024poseidon}, and Aurora adapts a large pretrained Earth-system model to multiple geophysical forecasting applications \citep{bodnar2025aurora}. A fusion-oriented perspective has further emphasized aligning foundation-model interfaces with experimental data and workflows \citep{churchill2025fusionfm}. Within tokamak modeling, TokaMind uses signal-aware tokens to support deterministic prediction across multiple forward tasks \citep{boschi2026tokamind}. From this perspective, GenToka focuses on a generative interface in which heterogeneous context, physical roles, and the requested target schema jointly define the output, allowing the same formulation to span forward prediction and the additional task families studied here.

\subsection{Diffusion for temporal and scientific generation}

Diffusion and score-based models represent complex conditional distributions through iterative denoising \citep{ho2020ddpm,song2021score}, and Transformer denoisers extend this principle to structured token sequences \citep{peebles2023dit}. Initial time-series applications established probabilistic forecasting and imputation with diffusion \citep{timegrad2021,csdi2021}. More recent methods combine diffusion with structured state-space models for multivariate forecasting and imputation \citep{alcaraz2023sssd}, incorporate temporal dynamics into spatiotemporal forecasting \citep{cachay2023dyffusion}, and reconcile next-token prediction with full-sequence diffusion \citep{chen2024diffusionforcing}. At the scale of scientific applications, GenCast demonstrates diffusion-based ensemble forecasting over global weather fields \citep{price2025gencast}, and FluidZero shows that a single generative model can master diverse tasks in fluid systems through shared conditional diffusion \citep{feng2025fluidzero}. These works establish diffusion as a versatile model for temporal uncertainty, high-dimensional physical evolution, and multi-task reuse within a scientific domain. Their interfaces, however, are generally built around a fixed feature tensor, spatial field, or sequence type. GenToka instead treats the requested signal set and its physical roles as part of the conditioning problem, and generates heterogeneous targets in their task-defined native structures.




\section{A Benchmark for Heterogeneous Tokamak Tasks}
\label{sec:benchmark}

The benchmark extends TokaMark \citep{rousseau2026tokamark}: it retains the fourteen forward tasks in Groups~1--4 and adds six tasks in Groups~5--6 for actuator action inversion and disruption prediction, yielding twenty MAST tasks that jointly exercise plasma-response modeling, actuator recovery, and disruption risk. Every task is defined by a fixed schema---permitted observations, requested outputs, temporal support, and evaluation cohort---so that models are compared on the same physical quantities. Section~\ref{sec:benchmark-protocol} states the shared evaluation rules; Section~\ref{sec:benchmark-reference-results} reports reference models on the full suite.

\subsection{Task Suite Overview}
\label{sec:benchmark-overview}

Table~\ref{tab:task-families} lists the six task groups. Groups~1--4 ask for reconstruction or forecasting of plasma responses under supplied operating context; Group~5 recovers recorded control-signal time series from an observed response in the same discharge; Group~6 estimates disruption risk from strictly pre-anchor history. Per-task signal lists and window lengths for the Group~5 and Group~6 schemas appear in Appendix~\ref{sec:appendix-task-schemas} (Table~\ref{tab:expanded-task-schemas}).

\begin{table}[t]
    \centering
    \small
    \setlength{\tabcolsep}{3pt}
    \renewcommand{\arraystretch}{1.12}
    \caption{Tokamak task families considered in this work. Each family can contain multiple context and target schemas without changing its underlying objective.}
    \label{tab:task-families}
    \begin{tabular}{@{}
        >{\raggedright\arraybackslash}p{0.25\linewidth}
        >{\raggedright\arraybackslash}p{0.34\linewidth}
        >{\raggedright\arraybackslash}p{0.33\linewidth}@{}}
        \toprule
        \textbf{Task family} & \textbf{Conditioning information} &
        \textbf{Requested output} \\
        \midrule
        Equilibrium reconstruction & Co-temporal magnetic and plasma observations &
        Global equilibrium quantities, plasma boundary, or poloidal-flux field \\
        Magnetic dynamics & Recent magnetic and plasma state with task-provided actuator traces &
        Future magnetic and equilibrium time series \\
        Profile and diagnostic dynamics & Recent profiles and diagnostics with operating context &
        Future kinetic profiles, global plasma quantities, or diagnostic responses \\
        Long-horizon MHD dynamics & Extended multimodal history with operating context &
        Future plasma-current, equilibrium, or diagnostic time series \\
        Actuator action inversion & State and control history with an observed
        same-discharge future response & Future recorded control-signal time series \\
        Disruption prediction & Strictly pre-anchor diagnostic and, where configured, control-related
        history & A scalar time-to-disruption risk surrogate \\
        \bottomrule
    \end{tabular}
\end{table}

\subsection{Task Schema Formalism}
\label{sec:problem-statement}

Tokamak signals differ in shape, and their roles depend on the modeling objective. A task schema~$\tau$, anchored at physical time~$t_c$, specifies which measurements enter as context and which quantities are requested as targets. The context component $\mathcal{S}_{\mathcal{C}}^{\tau}$ and target component $\mathcal{S}_{\mathcal{T}}^{\tau}$ enumerate role-qualified entries through index sets $\mathcal{I}_{\mathcal{C}}^{\tau}$ and $\mathcal{I}_{\mathcal{T}}^{\tau}$. Each entry~$k$ carries a descriptor $\varsigma_k$ with signal identity, modality, physical role, temporal support relative to~$t_c$, native shape, and availability semantics. Schema flexibility means reusing the same underlying signals under different role assignments and time intervals through task-specific adaptation.

We make the following notations: $X_\tau$ for diagnostic and plasma-state signals, $U_\tau$ for actuator or operating signals, $R_\tau$ for plasma responses, and $A_\tau$ for recorded control signals targeted in actuator action inversion. Superscripts $-$, $\mathrm{co}$, and $+$ mark support before, co-temporal with, and after~$t_c$; the superscript $\mathrm{ctx}$ marks quantities routed through the conditioning pathway, including any schema-admitted post-anchor operating segment.

\subsection{Plasma-Response Tasks}
\label{sec:benchmark-plasma-response}

Groups~1--4 evaluate reconstruction or forecasting of plasma responses. The model conditions on diagnostic and state observations $X_{\tau}^{\mathrm{ctx}}$ together with task-provided actuator information $U_{\tau}^{\mathrm{ctx}}$. Reconstruction requests the co-temporal response $R_{\tau}^{\mathrm{co}}$; forecasting requests the future response $R_{\tau}^{+}$, possibly with actuator context that extends into the prediction interval. These tasks therefore measure how accurately a model predicts plasma evolution given the observations and operating signals supplied by the schema.

\subsection{Actuator Action Inversion}
\label{sec:benchmark-actuator-inversion}

Group~5 is an inverse problem that reverses the usual input--output direction. Past state and control histories $(X_{\tau}^{-},U_{\tau}^{-})$ are combined with a future plasma response $R_{\tau,\mathrm{obs}}^{+}$ observed in the same discharge, and the model reconstructs the associated future control signals $A_{\tau}^{+}$. The targets are recorded PF-coil or gas-valve voltage traces. This group tests whether a model can recover the actuation compatible with an observed plasma response.

\subsection{Disruption Prediction}
\label{sec:benchmark-disruption}

Group~6 predicts a graded disruption-risk score from measurements strictly before~$t_c$. For a disruptive discharge with annotated event time $t_d>t_c$, remaining time $\Delta=t_d-t_c$ is mapped to a soft time-to-disruption (Soft-TTD) target $Z_{\tau,t_c}=\phi(\Delta)\in[0,1]$:
\begin{equation}
    \phi(\Delta)=
    \begin{cases}
        1, & \Delta\leq T_{\mathrm{urgent}},\\
        \left[1+\exp\!\left((\Delta-T_{\mathrm{mid}})/\lambda_{\mathrm{TTD}}\right)\right]^{-1},
            & T_{\mathrm{urgent}}<\Delta\leq T_{\mathrm{safe}},\\
        0, & \Delta>T_{\mathrm{safe}},
    \end{cases}
    \label{eq:soft-ttd}
\end{equation}
with urgent, midpoint, and safe time boundaries $T_{\mathrm{urgent}}<T_{\mathrm{mid}}<T_{\mathrm{safe}}$ and width $\lambda_{\mathrm{TTD}}>0$. 

\section{GenToka: A Generative Tokamak Foundation Model}
\label{sec:methodology}

GenToka builds on TokaMind \citep{boschi2026tokamind} and extends its supervised interface to a conditional generative formulation. Like TokaMind, heterogeneous measurements are role-qualified and encoded as tokens with signal-specific adapters; unlike TokaMind, each requested output is represented as a noisy target and modeled jointly through shared conditional diffusion rather than deterministic point regression. Signal-specific encoders and decoders still handle the geometry and temporal support of each measurement, while a diffusion Transformer captures dependencies among observations and requested outputs. We describe the conditional formulation, signal representations, joint denoising, and native-space reconstruction, followed by foundation pretraining and task-specific adaptation.

\subsection{Jointly Conditional Formulation}

Despite their different physical objectives, these tasks share the same computational structure: heterogeneous observations establish a condition, and the schema selects a variable set of signals to be generated jointly. For one example, the realized context collection is $\mathcal{C}_{\tau}=\{(\mathbf{x}_i,\mathbf{m}_i):i\in \mathcal{I}_{\mathcal{C}}^{\tau}\}$, where $\mathbf{x}_i$ contains the native values of context entry $i$ and the same-shaped binary mask $\mathbf{m}_i$ marks values that are both observed and quality-valid. The requested values form $\mathcal{Y}_{\tau}=\{\mathbf{y}_j:j\in\mathcal{I}_{\mathcal{T}}^{\tau}\}$. These arrays may differ in sampling grid, dimensionality, and native output space; for actuator action inversion, the paired future response enters $\mathcal{C}_{\tau}$ with its observed-response role.

The common schema-conditioned objective is
\begin{equation}
    p_{\theta_{\tau}}\!\left(
        \mathcal{Y}_{\tau}
        \mid \mathcal{C}_{\tau},
        \mathcal{S}_{\mathcal{C}}^{\tau},\mathcal{S}_{\mathcal{T}}^{\tau}
    \right).
    \label{eq:unified-task}
\end{equation}
The target collection represents a plasma response in Groups 1--4, recorded control signals in Group~5, and the risk surrogate in Group~6. GenToka models this collection jointly to capture dependencies among the requested signals. The task-specific parameters $\theta_{\tau}$ are initialized from shared foundation parameters $\theta_{\mathrm{F}}$ and then specialized to the corresponding context and target schema. In every downstream schema, a component-time element requested for generation is excluded from all conditioning paths, and unavailable or invalid target values do not contribute supervision.

\subsection{Overview of GenToka}
\label{sec:gentoka-overview}

GenToka generates the output signals requested by a task from its available measurements. The observed context remains clean, and diffusion is applied only to the target signals. This shared computation supports all six task groups: each task specifies the observed inputs and requested outputs, and the model learns their conditional relationship.

The schema first selects the available signals and assigns their physical and temporal roles. Signal histories are partitioned in physical time and compressed into context tokens, while each requested target window or scalar is represented by one noisy target token. A shared Transformer places all context and target tokens in the same self-attention sequence, allowing a target to draw on the complete schema-provided context and on the other requested signals. Iterative denoising then turns the target tokens into clean signal representations, which signal-specific decoders restore to time series, profiles, fields, or scalars in their native spaces.

This division of labor is central to the design. Deterministic codecs and lightweight adapters handle the geometry and scale that cannot be shared across signals, whereas the Transformer learns the dependencies that recur across schemas. One token represents a complete requested target, whether a window or a scalar, so the target-token count equals the number of requested signals and is independent of their sampling density.

\begin{figure}[t]
    \centering
    \includegraphics[width=\linewidth]{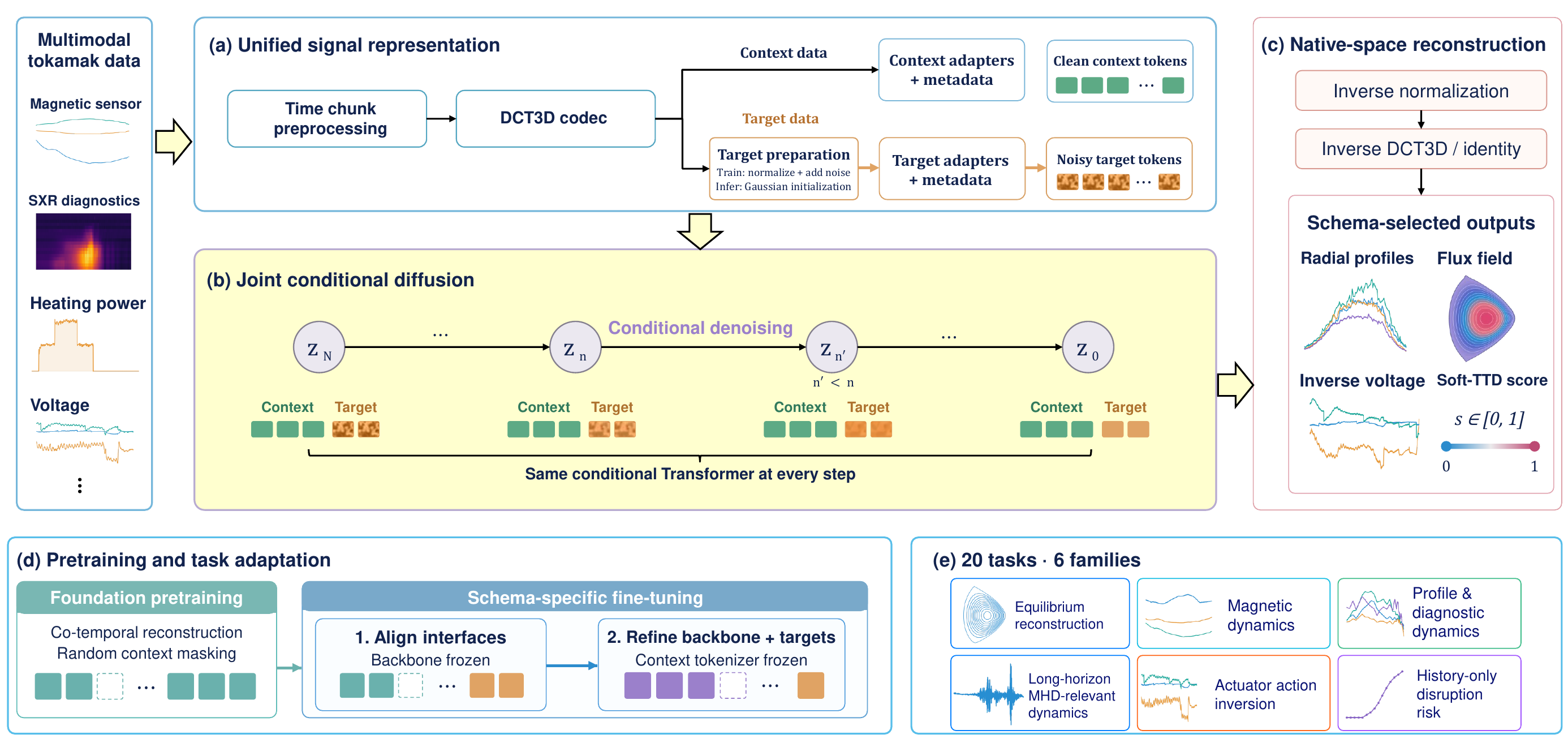}
    \caption{Overview of GenToka. (a)~Unified signal representation: multimodal measurements
    are time-chunked and encoded with signal-specific DCT3D; context and
    target adapters with schema metadata yield clean context tokens and noisy target tokens.
    (b)~Joint conditional diffusion: the same conditional Transformer denoises all requested
    targets jointly at each noise level while context tokens remain clean. (c)~Native-space
    reconstruction: denoised target codes are inverse-normalized and decoded to
    schema-selected time series, profiles, fields, or scalar Soft-TTD scores. (d)~Pretraining
    and task adaptation: foundation training uses co-temporal reconstruction with random
    context masking; each downstream schema is adapted in two stages by aligning interfaces
    with the backbone frozen, then refining the backbone and target pathway with the context
    tokenizer frozen. (e)~Benchmark scope: twenty MAST tasks in six families: equilibrium
    reconstruction, magnetic dynamics, profile and diagnostic dynamics, long-horizon MHD-relevant dynamics, actuator action inversion, and disruption prediction.}
    \label{fig:gentoka-overview}
\end{figure}

Figure~\ref{fig:gentoka-overview} summarizes this computation from the heterogeneous observations
to the native-space outputs. The schema determines which measurements provide clean context and
which signals are generated, while the same target-token pathway supports time series, profiles,
fields, and scalar risk scores. The lower panels place this inference procedure in the training
workflow: foundation pretraining learns reusable cross-signal representations, and task adaptation
then refines the interfaces and target pathway for each schema.

\subsection{Architecture of GenToka}
\label{sec:gentoka-architecture}

A shared generative model for tokamak data faces two coupled sources of heterogeneity: native signal geometry and task-dependent signal selection. GenToka handles them with signal-specific deterministic codecs and schema-selected token adapters, leaving the shared diffusion Transformer to model statistical dependencies across signals. The architecture proceeds from signal representation to token construction, target-space diffusion, joint denoising, and reconstruction.

\subsubsection{Heterogeneous Signal Representation and DCT3D Codec}

Tokamak diagnostics are sampled at markedly different rates, so direct concatenation into a synchronous vector requires unnecessary resampling and can obscure native temporal resolution. GenToka instead partitions each context signal into aligned intervals of fixed physical duration while retaining its native sampling grid. A signal--interval pair produces one context code; the most recent retained interval has relative position $p=1$, followed by progressively earlier intervals. A complete requested target is encoded once, preserving its internal structure when it is a window without creating one target token per sample.

For context entry $i$, let $\mathbf{x}_{i,p}$ be the native-valued chunk at relative position $p$. The fixed preprocessing map $\mathcal{T}_i$ produces $\widetilde{\mathbf{x}}_{i,p}=\mathcal{T}_i(\mathbf{x}_{i,p})$, and the reshape operator $\mathcal{A}_i$ views a scalar trace, profile, or time-dependent field as a tensor of shape $(1,1,T_i)$, $(H_i,1,T_i)$, or $(H_i,W_i,T_i)$, respectively, where $H_i$, $W_i$, and $T_i$ are the positive spatial and temporal axis lengths of that representation. GenToka then applies an orthonormal three-dimensional discrete cosine transform (DCT3D). With $\times_d$ denoting matrix multiplication along tensor axis $d$, the transform and coefficient extraction are
\begin{equation}
\begin{aligned}
    (\mathbf{D}_{L})_{k,\ell}
    &=\sqrt{\frac{2-\delta_{k0}}{L}}
      \cos\!\left[\frac{\pi}{L}\left(\ell+\frac{1}{2}\right)k\right],
      \quad 0\leq k,\ell<L,\\
    \mathbf{C}_{i,p}
    &=\mathcal{D}_i\!\left(\mathcal{A}_i(\widetilde{\mathbf{x}}_{i,p})\right)\\
    &:=\mathcal{A}_i(\widetilde{\mathbf{x}}_{i,p})
      \times_1\mathbf{D}_{H_i}
      \times_2\mathbf{D}_{W_i}
      \times_3\mathbf{D}_{T_i},
    \qquad
    \mathbf{c}_{i,p}^{\mathrm{ctx}}
    =\mathbf{S}_{\Omega_i}\operatorname{vec}(\mathbf{C}_{i,p}).
\end{aligned}
    \label{eq:dct3d-codec}
\end{equation}
Here $\mathbf{D}_L\in\mathbb{R}^{L\times L}$ is the orthonormal DCT-II matrix for an axis of length $L$; $k$ and $\ell$ index its frequency and sample coordinates, and $\delta_{k0}$ is the Kronecker delta. Writing $Q_i=H_iW_iT_i$, the set $\Omega_i\subseteq\{1,\ldots,Q_i\}$ contains $K_i$ retained flattened coordinates, and $\mathbf{S}_{\Omega_i}\in\{0,1\}^{K_i\times Q_i}$ is the matrix that selects them. Thus, $\mathbf{C}_{i,p}$ is the full coefficient tensor and $\mathbf{c}_{i,p}^{\mathrm{ctx}}\in\mathbb{R}^{K_i}$ is its retained context code. Singleton spatial axes allow the same operation to cover traces, profiles, and fields.

The same role-qualified codec applied to the complete target value $\mathbf{y}_j$ produces the clean target code $\mathbf{c}_{j}^{\mathrm{tar}}\in\mathbb{R}^{K_j}$, using its own preprocessing, geometry, and support $\Omega_j$. We write $\widetilde{\mathbf{y}}_j=\mathcal{T}_j(\mathbf{y}_j)$ for the corresponding preprocessed target. For decoded targets, $\mathcal{T}_j$ is paired with the fixed inverse preprocessing map $\mathcal{T}_j^{-1}$.

Each signal has a retained coefficient support selected from its training-set energy statistics. Its training-set energy statistics and selected indices are
\begin{equation}
    e_{i,q}=\mathbb{E}_{\mathbf{x}_{i,p}\sim\mathbb{P}_{i,\mathrm{train}}}
    \!\left[\bigl[\operatorname{vec}(\mathbf{C}_{i,p})\bigr]_{q}^{2}\right],
    \qquad
    \Omega_i\in\underset{
        \substack{\Omega\subseteq\{1,\ldots,Q_i\}\\|\Omega|=K_i}
    }{\arg\max}
    \sum_{q\in\Omega} e_{i,q},
    \label{eq:dct3d-support}
\end{equation}
where $e_{i,q}$ is the training-set second moment, or energy, of flattened coefficient $q$ for entry $i$, and $\mathbb{P}_{i,\mathrm{train}}$ is the empirical distribution of its training chunks. The retained dimension $K_i$ is selected to meet a prescribed explained-energy target subject to modality-aware coverage constraints and a configured representation budget. Traces, profiles, and fields can distribute their energy differently across temporal and spatial frequencies, so each role-qualified signal receives a separate support. Once estimated from training data, the support remains fixed. Quantities for which compression is unnecessary pass through an identity codec.

\subsubsection{Schema-Aware Token Construction}

The codec resolves differences in native shape, but its output dimension remains signal dependent, so signal-specific adapters map each code into the common hidden dimension $D$. The context adapter $\rho_i:\mathbb{R}^{K_i}\rightarrow\mathbb{R}^{D}$ is linear, $\rho_i(\mathbf{c}_{i,p}^{\mathrm{ctx}}) =\mathbf{P}_i\mathbf{c}_{i,p}^{\mathrm{ctx}}+\mathbf{b}_i$, with $\mathbf{P}_i\in\mathbb{R}^{D\times K_i}$ and $\mathbf{b}_i\in\mathbb{R}^{D}$. It acts on the deterministic code of an observed chunk, whereas $\eta_j:\mathbb{R}^{K_j}\rightarrow\mathbb{R}^{D}$ maps the normalized, diffusion-corrupted target code $\mathbf{z}_{j,n}$. Context codes remain clean and unchanged across reverse steps, while target content changes with the diffusion level and the current reverse update.

For a schema descriptor $\varsigma_k$, let $\iota_k$, $\nu_k$, and $\chi_k$ denote its underlying signal identity, modality, and assigned role. The learned lookup maps $\operatorname{Emb}_{\mathrm{pos}}$, $\operatorname{Emb}_{\mathrm{sig}}$, $\operatorname{Emb}_{\mathrm{mod}}$, and $\operatorname{Emb}_{\mathrm{role}}$ all return vectors in $\mathbb{R}^{D}$; $p_{\mathrm{out}}$ and $\mathrm{out}$ are the shared target-position and output-role labels. The resulting tokens are
\begin{equation}
\begin{aligned}
    \mathbf{e}_{i,p}^{\mathrm{ctx}}
    &=\rho_i(\mathbf{c}_{i,p}^{\mathrm{ctx}})
      +\operatorname{Emb}_{\mathrm{pos}}(p)
      +\operatorname{Emb}_{\mathrm{sig}}(\iota_i)
      +\operatorname{Emb}_{\mathrm{mod}}(\nu_i)
      +\operatorname{Emb}_{\mathrm{role}}(\chi_i),\\
    \mathbf{e}_{j,n}^{\mathrm{tar}}
    &=\eta_j(\mathbf{z}_{j,n})
      +\operatorname{Emb}_{\mathrm{pos}}(p_{\mathrm{out}})
      +\operatorname{Emb}_{\mathrm{sig}}(\iota_j)
      +\operatorname{Emb}_{\mathrm{mod}}(\nu_j)
      +\operatorname{Emb}_{\mathrm{role}}(\mathrm{out}).
\end{aligned}
    \label{eq:schema-aware-tokens}
\end{equation}
Equation~\ref{eq:schema-aware-tokens} is where the paths meet, but their semantics remain distinct. Context tokens carry relative chunk position $p$ and a task-specific role; target tokens use the fixed position $p_{\mathrm{out}}$ and output role because their content spans the complete output object, either a window or a scalar. Both token types lie in $\mathbb{R}^{D}$. Here $p$ indexes physical-time chunks, whereas $n$ denotes diffusion level; the two axes are unrelated. Signal and modality embeddings identify the underlying quantity and native structure. The schema determines sequence composition, metadata, and adapter selection. Unavailable entries are excluded from the valid token set, while target validity controls supervision. The next subsection specifies the normalization and corruption that produce $\mathbf{z}_{j,n}$.

\subsubsection{Diffusion in Target Representation Space}

The target pathway above leaves two operations to specify: normalization and diffusion corruption. During training, the clean code $\mathbf{c}_{j}^{\mathrm{tar}}$ is first centered and scaled using statistics from target $j$ in the training partition, aligning heterogeneous targets before Gaussian corruption:
\begin{equation}
\begin{aligned}
    \boldsymbol{\mu}_j
    &=\mathbb{E}_{\mathbb{P}_{j,\mathrm{train}}}
      [\mathbf{c}_{j}^{\mathrm{tar}}],\\
    \widetilde{\sigma}_{j,k}^{2}
    &=\max\!\left\{
        \operatorname{Var}_{\mathbb{P}_{j,\mathrm{train}}}
        ([\mathbf{c}_{j}^{\mathrm{tar}}]_k),s_{\min}^{2}
      \right\},\\
    \kappa_j^2
    &=\frac{1}{K_j}\sum_{k=1}^{K_j}\widetilde{\sigma}_{j,k}^{2},
    \qquad
    \mathbf{z}_{j,0}
    =\frac{\mathbf{c}_{j}^{\mathrm{tar}}-\boldsymbol{\mu}_j}{\kappa_j}.
\end{aligned}
    \label{eq:diffusion-normalization}
\end{equation}
Here $\mathbb{P}_{j,\mathrm{train}}$ is the empirical distribution of clean codes for target $j$, $k\in\{1,\ldots,K_j\}$ indexes a retained coefficient, $\boldsymbol{\mu}_j\in\mathbb{R}^{K_j}$ is the coordinate-wise mean, $s_{\min}>0$ is a standard-deviation floor, and $\kappa_j>0$ is the single root-mean-variance scale shared by all coordinates of target $j$. The normalized clean state $\mathbf{z}_{j,0}\in\mathbb{R}^{K_j}$ uses diffusion index $0$ exclusively for uncorrupted content. This normalization prevents high-amplitude outputs from dominating a common objective while retaining relative scale within one target representation.

For an $N$-level variance-preserving schedule, the noisy-state index is $n\in\{1,\ldots,N\}$ and each variance increment satisfies $0<\beta_n<1$. We define $\bar{\alpha}_n=\prod_{\ell=1}^{n}(1-\beta_\ell)$, with $\bar{\alpha}_0=1$, and write $a_n=\sqrt{\bar{\alpha}_n}$ and $b_n=\sqrt{1-\bar{\alpha}_n}$. Forward corruption and the velocity target then form an orthogonal rotation, from which the clean code and noise can be recovered:
\begin{equation}
\begin{aligned}
    \begin{bmatrix}
        \mathbf{z}_{j,n}\\[1pt]
        \mathbf{v}_{j,n}
    \end{bmatrix}
    &=
    \begin{bmatrix}
        a_n & b_n\\
        -b_n & a_n
    \end{bmatrix}
    \begin{bmatrix}
        \mathbf{z}_{j,0}\\[1pt]
        \boldsymbol{\epsilon}_j
    \end{bmatrix},
    \qquad \boldsymbol{\epsilon}_j\sim\mathcal{N}(\mathbf{0},\mathbf{I}_{K_j}),\\
    \widehat{\mathbf{z}}_{j,0}
    &=a_n\mathbf{z}_{j,n}-b_n\widehat{\mathbf{v}}_{j,n},
    \qquad
    \widehat{\boldsymbol{\epsilon}}_{j,n}
    =b_n\mathbf{z}_{j,n}+a_n\widehat{\mathbf{v}}_{j,n}.
\end{aligned}
    \label{eq:target-space-diffusion}
\end{equation}
This construction samples $\mathbf{z}_{j,n}$ from a Gaussian distribution centered at $a_n\mathbf{z}_{j,0}$ with covariance $b_n^2\mathbf{I}_{K_j}$. The vector $\boldsymbol{\epsilon}_j$ is the independent standard Gaussian perturbation of target $j$, $\mathbf{z}_{j,n}$ its corrupted state, and $\mathbf{v}_{j,n}$ the corresponding velocity target; hats denote model-derived estimates. All targets in an example share $n$, so the joint denoiser sees them at the same schedule-defined noise level, while their perturbations remain independent. Throughout this process, context tokens remain clean and provide a stable reference from which increasingly corrupted targets can be recovered. Small diffusion levels retain much of the target structure; large levels require the model to reconstruct it primarily from the schema and observed context.

\subsubsection{Joint Denoising Transformer}

At a fixed noise level, denoising is not a collection of independent per-signal problems. A plausible value for one target can depend on both the heterogeneous context and the other requested targets. Let $J_{\tau}=|\mathcal{I}_{\mathcal{T}}^{\tau}|$ be the number of requested target entries and $L_{\tau}$ the number of valid context-chunk tokens after temporal partitioning. The schema order induces a bijection $\pi_\tau:\mathcal{I}_{\mathcal{T}}^{\tau}\rightarrow\{1,\ldots,J_\tau\}$ that assigns each target entry its row in the target-token stack. Stacking the tokens from Equation~\ref{eq:schema-aware-tokens} gives $\mathbf{E}_{n}^{\mathrm{tar}}\in\mathbb{R}^{J_\tau\times D}$ and $\mathbf{E}^{\mathrm{ctx}}\in\mathbb{R}^{L_\tau\times D}$. GenToka places targets first and concatenates the stacks along the token axis:
\begin{equation}
    \mathbf{H}_{n}^{(0)}=
    [\mathbf{E}_{n}^{\mathrm{tar}}\Vert\mathbf{E}^{\mathrm{ctx}}]
    \in\mathbb{R}^{(J_{\tau}+L_{\tau})\times D}.
    \label{eq:joint-token-sequence}
\end{equation}
Full multi-head self-attention over this sequence creates direct context-to-target and target-to-target information paths, so independent Gaussian corruption does not imply an independent output distribution.

The diffusion step is embedded with sinusoidal features and a multilayer perceptron to obtain the global condition $\mathbf{d}_n$. Following diffusion Transformers \citep{peebles2023dit,ma2025latte}, every layer receives this condition through adaptive normalization. For attention ($u=A$) and feed-forward ($u=F$) sublayers of Transformer block $\ell\in\{0,\ldots,N_{\mathrm{blk}}-1\}$, the generic argument $\mathbf{H}$ below denotes a token matrix with hidden width $D$:
\begin{equation}
\begin{aligned}
    \operatorname{AdaLN}_{\ell}^{u}(\mathbf{H};\mathbf{d}_n)
    &=\bigl(1+\mathbf{s}_{\ell}^{u}(\mathbf{d}_n)\bigr)\odot
      \operatorname{LN}_{\ell}^{u}(\mathbf{H})
      +\boldsymbol{\delta}_{\ell}^{u}(\mathbf{d}_n),\\
    \mathbf{U}_{n}^{(\ell)}
    &=\mathbf{H}_{n}^{(\ell)}
      +\mathbf{g}_{\ell}^{A}(\mathbf{d}_n)\odot
      \operatorname{MSA}_{\ell}\!\left(
          \operatorname{AdaLN}_{\ell}^{A}(\mathbf{H}_{n}^{(\ell)};\mathbf{d}_n);
          M_{\mathrm{tok}}
      \right),\\
    \mathbf{H}_{n}^{(\ell+1)}
    &=\mathbf{U}_{n}^{(\ell)}
      +\mathbf{g}_{\ell}^{F}(\mathbf{d}_n)\odot
      \operatorname{MLP}_{\ell}\!\left(
          \operatorname{AdaLN}_{\ell}^{F}(\mathbf{U}_{n}^{(\ell)};\mathbf{d}_n)
      \right),
\end{aligned}
    \label{eq:gentoka-adaln-block}
\end{equation}
Here $\mathbf{d}_n\in\mathbb{R}^{D}$ is the learned embedding of noise level $n$; $\mathbf{s}_{\ell}^{u}(\mathbf{d}_n)$, $\boldsymbol{\delta}_{\ell}^{u}(\mathbf{d}_n)$, and $\mathbf{g}_{\ell}^{u}(\mathbf{d}_n)$ are its learned scale, shift, and residual-gate vectors in $\mathbb{R}^{D}$. Layer normalization $\operatorname{LN}_{\ell}^{u}$ acts on each token, $\odot$ denotes feature-wise multiplication broadcast over tokens, and $\operatorname{MSA}_{\ell}$ and $\operatorname{MLP}_{\ell}$ are the block-specific multi-head self-attention and feed-forward networks. The binary mask $M_{\mathrm{tok}}\in\{0,1\}^{J_\tau+L_\tau}$ equals one at valid token positions and prevents zero-valued padding positions from participating in attention; $\mathbf{U}_{n}^{(\ell)}$ is the intermediate state after the attention residual update. Zero-initializing the modulation layer makes each residual block initially an identity map.

After $N_{\mathrm{blk}}$ Transformer blocks, final layer normalization gives $\overline{\mathbf{H}}_n=\operatorname{LN}_{\mathrm{out}} (\mathbf{H}_n^{(N_{\mathrm{blk}})})$. Because the target tokens occupy the first $J_\tau$ rows, the contextualized representation supplied to the head of target $j$ is $\mathbf{h}_{j,n}=[\overline{\mathbf{H}}_n]_{\pi_\tau(j),:}\in\mathbb{R}^{D}$. This explicit row selection connects the shared Transformer state to each signal-specific prediction pathway.

\subsubsection{Signal-Specific Reconstruction}

Self-attention produces contextualized representations in a shared hidden space, but the requested signals still occupy different target-code spaces. Only hidden states at target positions are decoded. For target $j$, a compact head $\psi_j:\mathbb{R}^{D}\rightarrow\mathbb{R}^{K_j}$ predicts $\widehat{\mathbf{v}}_{j,n}=\psi_j(\mathbf{h}_{j,n})$; Equation~\ref{eq:target-space-diffusion} then supplies the corresponding clean-code estimate during reverse sampling. After the final update, inverse normalization and the deterministic codec restore the signal. For a coefficient tensor $\mathbf{C}\in\mathbb{R}^{H_j\times W_j\times T_j}$, orthonormality of the axis transforms gives the separable inverse $\mathcal{D}_j^{-1}(\mathbf{C})=\mathbf{C}\times_1\mathbf{D}_{H_j}^{\top} \times_2\mathbf{D}_{W_j}^{\top}\times_3\mathbf{D}_{T_j}^{\top}$. The complete reconstruction is
\begin{equation}
\begin{aligned}
    \widehat{\mathbf{c}}_{j}^{\mathrm{tar}}
    &=\boldsymbol{\mu}_j+\kappa_j\widehat{\mathbf{z}}_{j,0},\\
    \widehat{\widetilde{\mathbf{y}}}_{j}
    &=\mathcal{A}_{j}^{-1}\!\left[
        \mathcal{D}_{j}^{-1}\!\left(
            \operatorname{unvec}_{H_j,W_j,T_j}\!\left(
                \mathbf{S}_{\Omega_j}^{\top}
                \widehat{\mathbf{c}}_{j}^{\mathrm{tar}}
            \right)
        \right)
      \right],
    \qquad
    \widehat{\mathbf{y}}_j
    =\mathcal{T}_j^{-1}(\widehat{\widetilde{\mathbf{y}}}_{j}).
\end{aligned}
    \label{eq:signal-specific-reconstruction}
\end{equation}
Here $\mathbf{S}_{\Omega_j}^{\top}$ scatters the retained code back into the full $Q_j=H_jW_jT_j$ coefficient vector, filling coordinates outside $\Omega_j$ with zero, and $\operatorname{unvec}_{H_j,W_j,T_j}$ restores its coefficient-tensor shape. The inverse reshape $\mathcal{A}_j^{-1}$ recovers the preprocessed native geometry $\widehat{\widetilde{\mathbf{y}}}_j$, after which $\mathcal{T}_j^{-1}$ restores physical signal units. The truncated encoder--decoder composition is the orthogonal projection onto the retained spectral subspace. Targets assigned an identity codec bypass the inverse DCT and are reshaped directly before inverse preprocessing. The shared backbone is thus responsible for cross-signal reasoning, while signal-specific maps handle dimensions and coordinate systems that cannot be shared.

\subsection{Pretraining and Fine-Tuning}
\label{sec:training-finetuning}

GenToka separates broad representation acquisition from specialization to a downstream task. The foundation stage learns reusable relationships from a wide collection of co-temporal tokamak signals; fine-tuning then instantiates a particular conditioning and generation contract. Both stages use the same target-space velocity objective, so task adaptation does not introduce a new prediction mechanism. What changes is the schema that selects context and targets, their temporal relationship, and the subset of parameters updated during optimization. Task-specific codec supports and target-normalization statistics are estimated only from the corresponding training partition and remain fixed thereafter. In the common objective below, $\vartheta$ denotes the parameter state being optimized: $\theta_{\mathrm{F}}$ during foundation pretraining and $\theta_\tau$ during schema-specific fine-tuning.

\subsubsection{Foundation Pretraining}

The foundation model is trained with a single broad, co-temporal reconstruction schema. Its clean context comprises diagnostic, plasma-state, and actuator signals, while the target collection spans both observed diagnostics and additional state or equilibrium quantities. Some diagnostics therefore appear simultaneously as context and reconstruction targets. This shared reconstruction objective trains the model to capture dependencies among diagnostics, plasma state, and actuation for transfer to downstream schemas. To discourage reliance on a fixed, complete set of contemporaneous observations, training randomly removes individual context tokens and aligned physical-time chunks. This exposes the model to varying patterns of partial evidence and encourages it to exploit correlations across signals and history.

Training follows the velocity parameterization in Equation~\ref{eq:target-space-diffusion}. For the $N$ noisy levels defined above, each example $r\in\{1,\ldots,B\}$ draws $n_r\sim\operatorname{Unif}\{1,\ldots,N\}$; this level is shared by all of its requested targets, whose Gaussian perturbations remain independent. In a minibatch $\mathcal{B}$ of size $B$, the indicator $\omega_{rj}\in\{0,1\}$ equals one when target entry $j$ provides a valid supervised code for example $r$, and zero otherwise. The quantity $K_j$ is its code dimension. The masked denoising objective is
\begin{equation}
\begin{aligned}
    \ell_{rj}
    &=\frac{1}{K_j}
      \left\|\widehat{\mathbf{v}}_{j,n_r}^{(r)}
      -\mathbf{v}_{j,n_r}^{(r)}\right\|_2^2,
      \qquad V_j=\sum_{r=1}^{B}\omega_{rj},\\
    \mathcal{L}_{\mathrm{den}}(\vartheta;\tau)
    &=\mathbb{E}\!\left[
      \frac{1}{|\mathcal{J}_{\mathcal{B}}|}
      \sum_{j\in\mathcal{J}_{\mathcal{B}}}
      \frac{1}{V_j}\sum_{r=1}^{B}\omega_{rj}\ell_{rj}
      \right],
      \qquad
      \mathcal{J}_{\mathcal{B}}
      =\{j\in\mathcal{I}_{\mathcal{T}}^{\tau}:V_j>0\}.
\end{aligned}
    \label{eq:masked-denoising-loss}
\end{equation}
Here the superscript $(r)$ identifies the minibatch example, $\ell_{rj}$ is the mean squared velocity error for one example--target pair, $V_j$ counts its valid examples in the minibatch, and $\mathcal{J}_{\mathcal{B}}$ contains the targets with $V_j>0$; training batches contain at least one such target. The expectation covers minibatch sampling, diffusion levels, Gaussian noise, and random context masking. Averaging first over code coordinates prevents a high-dimensional signal from dominating solely because of its representation size; averaging over valid examples and then over available targets gives each supervised signal equal influence within a minibatch. We use no additional per-signal or timestep weights. All token adapters, the shared denoising Transformer, and target heads are optimized jointly with AdamW, using linear warmup followed by cosine learning-rate decay. Numerical settings are reported with the experimental protocol.

\subsubsection{Schema-Specific Fine-Tuning}

Fine-tuning instantiates Equation~\ref{eq:unified-task} with the roles and temporal support of each downstream task. For compact notation, $\mathcal{H}_{\tau}^{-}=(X_{\tau}^{-},U_{\tau}^{-})$ collects available pre-anchor diagnostic, state, and control history, and $U_{\tau,\mathrm{given}}^{+}$ denotes any post-anchor actuator or operating segment supplied by the schema. With the fixed schema components suppressed, the task-specific conditionals are
\begin{equation}
\begin{aligned}
    p_{\theta_{\tau}}(R_{\tau}^{\mathrm{co}}\mid X_{\tau}^{\mathrm{co}})
    &\quad \text{co-temporal equilibrium reconstruction},\\
    p_{\theta_{\tau}}(R_{\tau}^{+}\mid
        \mathcal{H}_{\tau}^{-},U_{\tau,\mathrm{given}}^{+})
    &\quad \text{conditional dynamics forecasting},\\
    p_{\theta_{\tau}}(A_{\tau}^{+}\mid
        \mathcal{H}_{\tau}^{-},R_{\tau,\mathrm{obs}}^{+})
    &\quad \text{actuator action inversion},\\
    p_{\theta_{\tau}}(Z_{\tau,t_c}\mid\mathcal{H}_{\tau}^{-})
    &\quad \text{disruption prediction}.
\end{aligned}
    \label{eq:task-specific-conditionals}
\end{equation}

Each schema routes permitted observations to clean context and requested signals to the target pathway; absent signal groups are omitted. The conditioning and supervision rules in Section~\ref{sec:problem-statement}, including the Soft-TTD construction in Equation~\ref{eq:soft-ttd}, apply throughout adaptation. All schemas use Equation~\ref{eq:masked-denoising-loss}; adaptation changes the signal roles, temporal support, and target codecs while retaining the same denoising objective.

The shared denoising Transformer is initialized from the foundation model, together with each context projection whose role-qualified signal identity and representation dimension remain compatible. New context maps and all target-side noisy-code projections and denoising heads are initialized for the downstream schema. This fresh target pathway is intentional: the same physical quantity can acquire a different temporal support, codec, or semantic role across tasks, for which direct interface transfer would conflate distinct contracts. Target normalization is re-estimated from the downstream training partition, while the optimizer state and learning-rate schedule are initialized anew.

Adaptation proceeds in two stages. The first optimizes the schema-facing context and target modules while holding the pretrained denoising backbone fixed, allowing new interfaces to align with its representation without immediately perturbing reusable cross-signal structure. The second freezes the aligned context tokenizer and jointly refines the backbone and target pathway. Task-specific dependencies can thereby be incorporated without moving the input representation at the same time.

\subsection{Inference on Specific Tasks}
\label{sec:task-specific-inference}

At inference, the selected schema instantiates one conditional contract from Equation~\ref{eq:task-specific-conditionals} and fixes the context and requested target set; target values are not supplied to the model. Each target representation is instead initialized independently from standard Gaussian noise, while the same clean context collection is supplied at every reverse step. All targets follow a common descending noise grid and are processed together by the joint Transformer.

GenToka uses a deterministic implicit reverse process \citep{song2021ddim} on a possibly subsampled grid $N=n_G>\cdots>n_1\geq 1$, where $G$ is the number of selected model-evaluation levels. For every $j\in\mathcal{I}_{\mathcal{T}}^{\tau}$, sampling begins with $\mathbf{z}_{j,n_G}\sim\mathcal{N}(\mathbf{0},\mathbf{I}_{K_j})$. At a nonterminal grid position $g=G,\ldots,2$, one joint model evaluation predicts $\widehat{\mathbf{v}}_{j,n_g}$ for all requested targets, after which
\begin{equation}
\begin{aligned}
    \widehat{\mathbf{z}}_{j,0}^{(g)}
    &=a_{n_g}\mathbf{z}_{j,n_g}
      -b_{n_g}\widehat{\mathbf{v}}_{j,n_g},\\
    \widehat{\boldsymbol{\epsilon}}_{j,n_g}
    &=b_{n_g}\mathbf{z}_{j,n_g}
      +a_{n_g}\widehat{\mathbf{v}}_{j,n_g},\\
    \mathbf{z}_{j,n_{g-1}}
    &=a_{n_{g-1}}\widehat{\mathbf{z}}_{j,0}^{(g)}
      +b_{n_{g-1}}\widehat{\boldsymbol{\epsilon}}_{j,n_g},\\
    \widehat{\mathbf{z}}_{j,0}
    &=a_{n_1}\mathbf{z}_{j,n_1}
      -b_{n_1}\widehat{\mathbf{v}}_{j,n_1},
\end{aligned}
    \label{eq:gentoka-reverse-sampling}
\end{equation}
at the terminal evaluation
The first two lines are the velocity-to-clean-state and velocity-to-noise conversions from Equation~\ref{eq:target-space-diffusion}; the third advances every target to the next shared noise level, and the fourth returns the clean normalized code from the last noisy level $n_1$. At grid position $g=1$, this terminal prediction is obtained by one final joint model evaluation. The number and placement of evaluation levels affect computation but not the task interface and are specified in the experimental protocol.

The terminal codes are inverse-normalized and restored to native signal spaces through Equation~\ref{eq:signal-specific-reconstruction}. A new initial Gaussian noise yields another conditional generation. The same joint reverse process therefore generates equilibrium responses, future plasma-response windows, recorded control-signal time series, or scalar Soft-TTD scores, with their conditioning roles and native output spaces determined by the schema.

\section{Experimental Results}
\label{sec:experiments}

The experiments provide a systematic evaluation across all 20 benchmark tasks. First, complete LSTM and TokaMark-CNN baseline results characterize the benchmark across six task groups. Second, we evaluate GenToka against TokaMind and use paired GenToka--Scratch experiments to examine the contribution of foundation pretraining.

\subsection{Benchmark and Evaluation Protocol}
\label{sec:benchmark-protocol}
\label{sec:experimental-setup}

\paragraph{Data and task protocols.}
All experiments use MAST data distributed through FAIR-MAST and the signal-processing and evaluation conventions of TokaMark \citep{jackson2024fairmast,jackson2025opendata,rousseau2026tokamark}. The disruption prediction signal is manually annotated by experts. The benchmark contains 20 tasks in six groups (Groups 1--6). Data partitions are disjoint at the shot level; normalization and codec statistics use training data, and model selection uses validation data only. Availability and quality rules are identical within each primary TokaMind--GenToka comparison.

\paragraph{Baseline model protocol.}
The LSTM and TokaMark-CNN baseline experiments cover all 20 tasks. Both use the benchmark's signal roles, temporal supports, frozen shot partitions, and native-space evaluator, with their respective representations and optimization settings specified in Appendix~\ref{sec:appendix-benchmark-baselines}.

\paragraph{Metrics.}
For Groups 1--5, prediction error is measured by normalized root mean squared error (NRMSE) in the native signal space. We retain $j$ for a role-qualified target entry, consistent with Section~\ref{sec:problem-statement}. For task schema $\tau$, let $\mathcal{V}_\tau$ be its valid test shots and $\mathcal{I}_{\tau,s}^{\mathrm{eval}}\subseteq\mathcal{I}_{\mathcal{T}}^\tau$ the target entries evaluated in shot $s$. The valid windows of entry $j$ form $\mathcal{W}_{s,j}$, and $\mathcal{Q}_{s,j,w}$ indexes the native points of window $w$. At point $q$, $y_{s,j,w,q}$ and $\widehat{y}_{s,j,w,q}$ are the observed and predicted values, while $\sigma_j>0$ is the training-set standard deviation of that target. The evaluator first aggregates windows for each shot--target pair and then gives equal weight to the targets within a shot and to the shots within a task:
\begin{equation}
\begin{aligned}
    \operatorname{NRMSE}_{s,j}
    &=\frac{1}{\sigma_j}
      \left[
        \frac{1}{|\mathcal{W}_{s,j}|}
        \sum_{w\in\mathcal{W}_{s,j}}
        \frac{1}{|\mathcal{Q}_{s,j,w}|}
        \sum_{q\in\mathcal{Q}_{s,j,w}}
        (\widehat{y}_{s,j,w,q}-y_{s,j,w,q})^2
      \right]^{1/2},\\
    \operatorname{NRMSE}_{\tau}
    &=\frac{1}{|\mathcal{V}_{\tau}|}
      \sum_{s\in\mathcal{V}_{\tau}}
      \frac{1}{|\mathcal{I}_{\tau,s}^{\mathrm{eval}}|}
      \sum_{j\in\mathcal{I}_{\tau,s}^{\mathrm{eval}}}
      \operatorname{NRMSE}_{s,j}.
\end{aligned}
    \label{eq:evaluation-nrmse}
\end{equation}
Within each group, we additionally report macro-averages over tasks so that high-dimensional outputs do not dominate the summary. For disruption prediction, we report window-level area under the receiver operating characteristic curve (AUROC) and F1. Windows whose Soft-TTD target is at least $0.5$ are treated as positive. AUROC is computed from the continuous risk score, whereas F1 requires a fixed operating threshold. To avoid task- or method-specific operating points, we evaluate the candidate set $\Theta=\{0.1,0.2,\cdots,0.9\}$ on the six primary model--task pairs (TokaMind and GenToka on each of the three disruption schemas). If $p_{m,v}$ and $y_{m,v}$ denote the validation scores and binary labels for pair $m$, we select
\begin{equation}
    \tau^\star
    = \min \arg\max_{\tau\in\Theta}
      \frac{1}{6}\sum_{m=1}^{6}
      F_1\left(y_{m,v},\mathbf{1}\left[p_{m,v}\geq\tau\right]\right),
    \label{eq:disruption-threshold}
\end{equation}
where the minimum resolves ties in favor of the lower threshold. The selected threshold is frozen before any test-set metric is computed and without using test labels; Scratch, LSTM, and TokaMark-CNN models use the same $\tau^\star$ but do not participate in its selection. This protocol gives every model--task pair equal weight and keeps threshold selection separate from test evaluation.

\begin{table}[!htb]
  \caption{GenToka model results across all twenty tasks. Groups 1--5 report native-space NRMSE; Group 6 reports window-level AUROC and F1, with $\tau^\star=0.4$ for F1. Means are unweighted over tasks. Scratch uses the GenToka architecture with random initialization and the same downstream recipe for each task. Bold identifies the best displayed value, including ties.}
  \label{tab:benchmark-results}
  \centering
  \small
  \setlength{\tabcolsep}{4pt}
  \begin{tabular}{lllrrr}
      \toprule
      Group & Task & Metric & TokaMind & GenToka & Scratch \\
      \midrule
      Group 1 & 1-1 & NRMSE $\downarrow$ & 0.1363 & \textbf{0.0941} & 0.0991 \\
       & 1-2 & NRMSE $\downarrow$ & 0.0439 & \textbf{0.0412} & \textbf{0.0412} \\
       & 1-3 & NRMSE $\downarrow$ & 0.1200 & \textbf{0.1156} & 0.1184 \\
       & \textit{Group 1 avg} & NRMSE $\downarrow$ & 0.1001 & \textbf{0.0836} & 0.0862 \\
      \midrule
      Group 2 & 2-1 & NRMSE $\downarrow$ & 0.1289 & \textbf{0.0880} & 0.0931 \\
       & 2-2 & NRMSE $\downarrow$ & 0.0408 & \textbf{0.0370} & 0.0375 \\
       & 2-3 & NRMSE $\downarrow$ & 0.0866 & \textbf{0.0828} & 0.0862 \\
       & \textit{Group 2 avg} & NRMSE $\downarrow$ & 0.0854 & \textbf{0.0692} & 0.0723 \\
      \midrule
      Group 3 & 3-1 & NRMSE $\downarrow$ & 0.3477 & \textbf{0.3311} & 0.3465 \\
       & 3-2 & NRMSE $\downarrow$ & 0.2529 & \textbf{0.2522} & 0.2943 \\
       & 3-3 & NRMSE $\downarrow$ & 0.2415 & \textbf{0.2147} & 0.2255 \\
       & \textit{Group 3 avg} & NRMSE $\downarrow$ & 0.2807 & \textbf{0.2660} & 0.2888 \\
      \midrule
      Group 4 & 4-1 & NRMSE $\downarrow$ & \textbf{0.2122} & 0.2298 & 0.2582 \\
       & 4-2 & NRMSE $\downarrow$ & \textbf{0.2143} & 0.2302 & 0.2663 \\
       & 4-3 & NRMSE $\downarrow$ & 0.0781 & \textbf{0.0691} & 0.0789 \\
       & 4-4 & NRMSE $\downarrow$ & 0.3470 & 0.3116 & \textbf{0.3064} \\
       & 4-5 & NRMSE $\downarrow$ & \textbf{0.7377} & 0.7487 & 0.7525 \\
       & \textit{Group 4 avg} & NRMSE $\downarrow$ & \textbf{0.3179} & \textbf{0.3179} & 0.3325 \\
      \midrule
      \multicolumn{2}{l}{\textit{Groups 1--4 avg}} & NRMSE $\downarrow$ & 0.2134 & \textbf{0.2033} & 0.2146 \\
      \midrule
      Group 5 & 5-1 & NRMSE $\downarrow$ & 0.1386 & \textbf{0.0750} & 0.0822 \\
       & 5-2 & NRMSE $\downarrow$ & 0.3206 & \textbf{0.3065} & 0.3234 \\
       & 5-3 & NRMSE $\downarrow$ & 0.2503 & \textbf{0.2024} & 0.2335 \\
      \midrule
       & \textit{Group 5 avg} & NRMSE $\downarrow$ & 0.2365 & \textbf{0.1946} & 0.2130 \\
      \midrule
Group 6 & 6-1 & AUROC $\uparrow$ & 0.9281 & 0.9213 & \textbf{0.9314} \\
       &  & F1 $\uparrow$ & 0.6029 & \textbf{0.6297} & 0.6186 \\
       & 6-2 & AUROC $\uparrow$ & 0.9242 & 0.9111 & \textbf{0.9246} \\
       &  & F1 $\uparrow$ & 0.5923 & \textbf{0.6206} & 0.6151 \\
       & 6-3 & AUROC $\uparrow$ & 0.8683 & \textbf{0.8824} & 0.8728 \\
       &  & F1 $\uparrow$ & 0.3891 & \textbf{0.4674} & 0.4427 \\
       & \textit{Group 6 avg} & AUROC $\uparrow$ & 0.9069 & 0.9049 & \textbf{0.9096} \\
       &  & F1 $\uparrow$ & 0.5281 & \textbf{0.5726} & 0.5588 \\
      \bottomrule
  \end{tabular}
\end{table}

The selected threshold is $\tau^\star=0.4$, with mean validation F1 $0.5570$. The full validation scan and its constituent scores are reported in Appendix~\ref{sec:appendix-threshold-selection}.

\subsection{Benchmark Reference Results}
\label{sec:benchmark-reference-results}

The reference experiments examine how the benchmark distinguishes physical targets, input--output relationships, and available diagnostic information. The evaluation suite contains twenty tasks in six groups: Groups 1--4 comprise the fourteen forward tasks inherited from TokaMark \citep{rousseau2026tokamark}, and Groups 5--6 comprise six newly created tasks---three actuator action inversion tasks (Group~5) and three disruption prediction tasks (Group~6). We train an LSTM and TokaMark-CNN from random initialization on each task; Table~\ref{tab:benchmark-references} reports their task-level results and group means. Patterns shared by these two reference implementations illustrate the capabilities probed by the task design. Architectures and training settings are detailed in Appendix~\ref{sec:appendix-benchmark-baselines}.

\begin{table}[!htb]
    \caption{Benchmark reference results for all twenty tasks. LSTM and TokaMark-CNN are trained separately for each task from random initialization and evaluated with the common native-space metrics. Groups 1--4 contain the fourteen TokaMark forward tasks; Groups 5 and 6 contain the six newly created tasks (actuator action inversion and disruption prediction). Means give equal weight to tasks. Disruption F1 uses the frozen threshold $\tau^\star=0.4$. The two reference implementations use their respective representations and training recipes (Appendix~\ref{sec:appendix-benchmark-baselines}).}
    \label{tab:benchmark-references}
    \centering
    \small
    \setlength{\tabcolsep}{8pt}
    \begin{tabular}{lllrr}
        \toprule
        Group & Task & Metric & LSTM & TokaMark-CNN \\
        \midrule
        Group 1 & 1-1 & NRMSE $\downarrow$ & 0.1853 & 0.1491 \\
         & 1-2 & NRMSE $\downarrow$ & 0.0488 & 0.0408 \\
         & 1-3 & NRMSE $\downarrow$ & 0.1302 & 0.1354 \\
         & \textit{Group 1 avg} & NRMSE $\downarrow$ & 0.1215 & 0.1084 \\
        \midrule
        Group 2 & 2-1 & NRMSE $\downarrow$ & 0.1633 & 0.1376 \\
         & 2-2 & NRMSE $\downarrow$ & 0.0461 & 0.0482 \\
         & 2-3 & NRMSE $\downarrow$ & 0.1046 & 0.1155 \\
         & \textit{Group 2 avg} & NRMSE $\downarrow$ & 0.1047 & 0.1004 \\
        \midrule
        Group 3 & 3-1 & NRMSE $\downarrow$ & 0.3722 & 0.3675 \\
         & 3-2 & NRMSE $\downarrow$ & 0.2825 & 0.3307 \\
         & 3-3 & NRMSE $\downarrow$ & 0.2776 & 0.2687 \\
         & \textit{Group 3 avg} & NRMSE $\downarrow$ & 0.3108 & 0.3223 \\
        \midrule
        Group 4 & 4-1 & NRMSE $\downarrow$ & 0.2489 & 0.2557 \\
         & 4-2 & NRMSE $\downarrow$ & 0.2596 & 0.2649 \\
         & 4-3 & NRMSE $\downarrow$ & 0.0948 & 0.1137 \\
         & 4-4 & NRMSE $\downarrow$ & 0.3514 & 0.5254 \\
         & 4-5 & NRMSE $\downarrow$ & 0.7347 & 0.7428 \\
         & \textit{Group 4 avg} & NRMSE $\downarrow$ & 0.3379 & 0.3805 \\
        \midrule
        \multicolumn{2}{l}{\textit{Groups 1--4 avg}} & NRMSE $\downarrow$ & 0.2357 & 0.2497 \\
        \midrule
        Group 5 & 5-1 & NRMSE $\downarrow$ & 0.1431 & 0.2927 \\
         & 5-2 & NRMSE $\downarrow$ & 0.3731 & 0.4603 \\
         & 5-3 & NRMSE $\downarrow$ & 0.3410 & 0.4440 \\
         & \textit{Group 5 avg} & NRMSE $\downarrow$ & 0.2858 & 0.3990 \\
        \midrule
        Group 6 & 6-1 & AUROC $\uparrow$ & 0.9270 & 0.9060 \\
         &  & F1 $\uparrow$ & 0.5940 & 0.5683 \\
         & 6-2 & AUROC $\uparrow$ & 0.9186 & 0.9101 \\
         &  & F1 $\uparrow$ & 0.5768 & 0.5607 \\
         & 6-3 & AUROC $\uparrow$ & 0.8542 & 0.8793 \\
         &  & F1 $\uparrow$ & 0.3900 & 0.4539 \\
         & \textit{Group 6 avg} & AUROC $\uparrow$ & 0.8999 & 0.8985 \\
         &  & F1 $\uparrow$ & 0.5203 & 0.5276 \\
        \bottomrule
    \end{tabular}
\end{table}

\paragraph{Separating physical targets within Groups 1--4.}
Tasks in Groups 1--4 distinguish reconstruction of equilibrium geometry from prediction of kinetic and magnetic dynamics. Both reference models obtain NRMSE below 0.05 for plasma-boundary reconstruction and forecasting (Tasks 1-2 and 2-2), while the profile and diagnostic family has mean errors above 0.31. A more targeted contrast appears within the long-horizon family: Tasks 4-3 and 4-5 use the same diagnostic and operating inputs and a 100\,ms prediction horizon, but request the vertical magnetic-axis position and Mirnov diagnostic signals, respectively. Their NRMSE ranges are 0.0948--0.1137 and 0.7347--0.7428 across the two reference models. Preserving this distinction exposes different modeling demands across macroscopic plasma position and magnetic fluctuations that a single aggregate forecasting score would obscure.

\paragraph{Evaluating the response-to-control relationship.}
Group~5 assesses actuator action inversion: recovering the recorded control signals associated with an observed plasma response. Task 5-1 reconstructs PF-coil voltages from current responses, whereas Tasks 5-2 and 5-3 reconstruct the same six gas-valve voltage channels from an electron-density profile and D-alpha signals, respectively. Both reference implementations yield lower errors for PF-voltage reconstruction than for gas-valve reconstruction, and distinguish the two gas-valve task configurations. They motivate retaining separate schemas for magnetic actuation and fueling, with their histories and response measurements. 

\paragraph{Assessing prediction under different diagnostic coverage.}
Group~6 holds the test windows and targets fixed while varying the historical inputs from nine historical signals to five and then to plasma current alone. With five input signals, F1 is 0.5768 for the LSTM and 0.5607 for TokaMark-CNN; with plasma current alone, the corresponding values are 0.3900 and 0.4539. Both references therefore distinguish the multivariate and current-only configurations, illustrating the value of evaluating diagnostic availability explicitly. AUROC and F1 further separate risk ranking from classification at the common operating threshold. This task family adds an event-oriented objective to the continuous-output benchmark and supports evaluation across different diagnostic configurations.

\subsection{GenToka Evaluation}
\label{sec:benchmark-results}

We compare GenToka directly with TokaMind \citep{boschi2026tokamind}. Both methods share the task schema, physical signal roles, shot split, and native-space evaluator within each primary comparison. GenToka uses signal-specific spectral codecs and token adapters instantiated from the task schema; one generated realization is evaluated for each test example. The two complete pipelines use their respective training budgets, prediction objectives, checkpoint-selection criteria, and generation procedures. Scratch uses the GenToka architecture and the same downstream recipe for each task, initialized randomly (Appendix~\ref{sec:appendix-model-protocols}). Table~\ref{tab:benchmark-results} organizes the comparison across all twenty tasks.

Across Groups 1--4, GenToka lowers NRMSE relative to TokaMind on eleven of fourteen tasks, reducing the Groups~1--4 macro-average from 0.2134 to 0.2033 (4.7\%).

\subsubsection{Overview}
\label{sec:initialization-results}

Table~\ref{tab:benchmark-results} reports the full task matrix. Relative to TokaMind, GenToka improves NRMSE on eleven of fourteen plasma-response tasks and lowers the Groups~1--4 macro-average by 4.7\%. It also reduces Group~5 NRMSE from 0.2365 to 0.1946 (17.7\%) and raises Group~6 macro-average F1 from 0.5281 to 0.5726 (8.4\%) at the shared threshold $\tau^\star=0.4$, while window-level AUROC remains comparable (0.9069 versus 0.9049). Scratch is the same GenToka architecture and downstream recipe trained from random initialization (Appendix~\ref{sec:appendix-model-protocols}). Pretrained GenToka lowers NRMSE relative to Scratch on twelve of fourteen Group~1--4 tasks and reduces the Groups~1--4 macro-average from 0.2146 to 0.2033 (5.3\%); Scratch matches GenToka on Task~1-2 and is slightly better on Task~4-4. On Group~5, pretraining lowers the group mean from 0.2130 to 0.1946 (8.6\%), and on Group~6 it improves macro-average F1 from 0.5588 to 0.5726. Together, the TokaMind and Scratch comparisons indicate that both the generative modeling interface and foundation pretraining contribute to the gains summarized in Table~\ref{tab:benchmark-results}. The subsections below walk through each task group with qualitative examples.

\subsubsection{Equilibrium Reconstruction}
\label{sec:equilibrium-results}
This group tests cross-signal reconstruction from co-temporal evidence. GenToka lowers the group-mean NRMSE from 0.1001 to 0.0836 relative to TokaMind (16.4\%), with lower error on all three tasks.
The same architecture trained from scratch averages 0.0862 NRMSE; foundation pretraining reduces this group mean to 0.0836 and improves Tasks~1-1 and 1-3, with a tie on Task~1-2, indicating that pretraining can improve the accuracy of equilibrium reconstruction.
\begin{figure}[!htpb]
    \centering
    \includegraphics[page=4,width=0.9\linewidth,keepaspectratio]{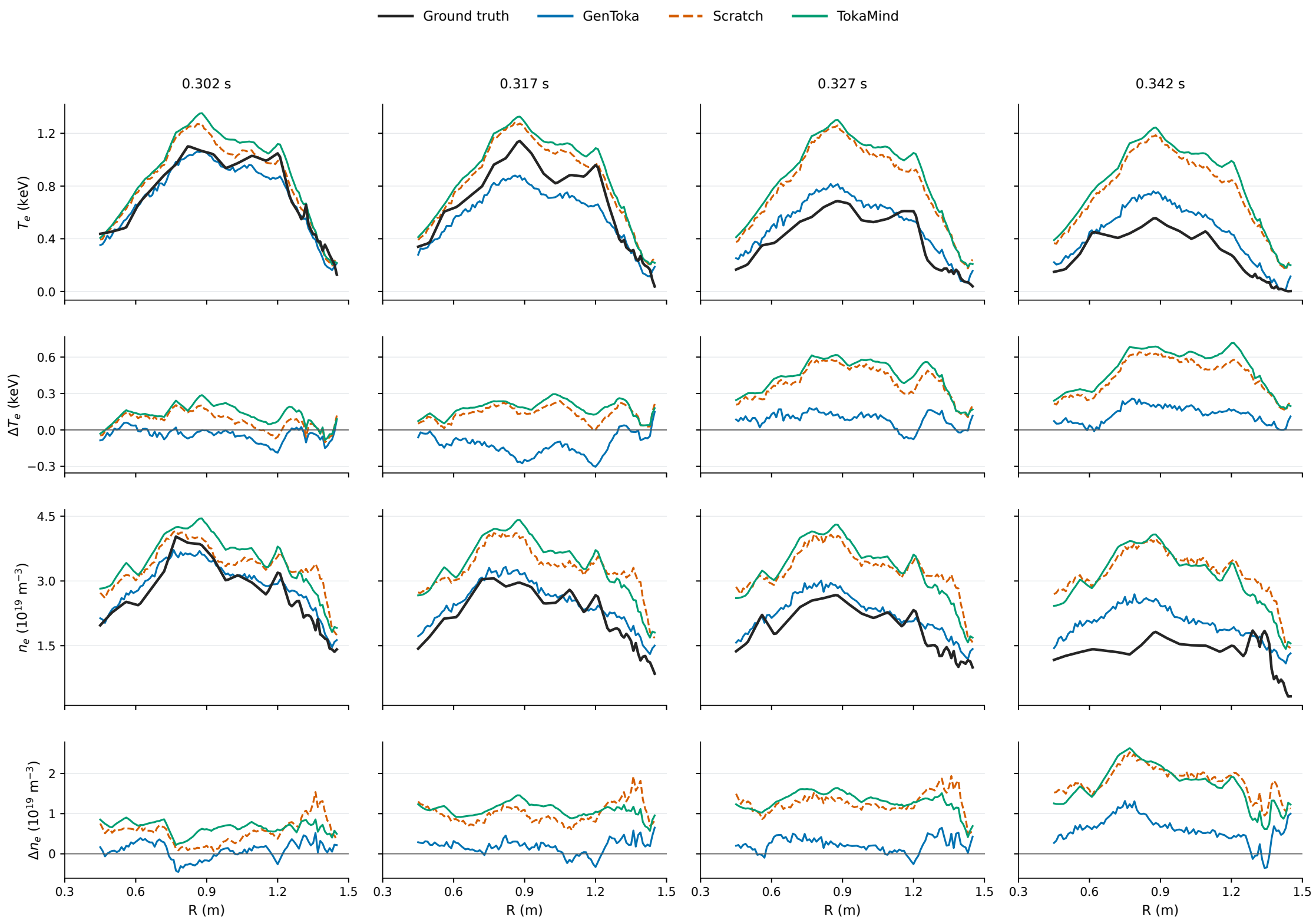}
    \caption{On a co-temporal reconstruction example, GenToka tracks the measured electron temperature and density profiles more closely than TokaMind, matching the lower aggregate equilibrium error summarized above.}
    \label{fig:qual-equilibrium}
\end{figure}

\subsubsection{Magnetic Dynamics}
\label{sec:magnetic-results}
The magnetic-dynamics family tests future response prediction under task-provided operating context. GenToka lowers NRMSE relative to TokaMind on all three tasks. The TokaMind group mean decreases from 0.0854 to 0.0692 (19.0\%), with the largest task-level reduction on Task 2-1, from 0.1289 to 0.0880.
Scratch reaches a 0.0723 group mean on these schemas; pretraining lowers NRMSE to 0.0692 (4.3\%) and outperforms Scratch on every magnetic-dynamics task.
\begin{figure}[!htpb]
    \centering
    \includegraphics[page=8,width=0.9\linewidth,keepaspectratio]{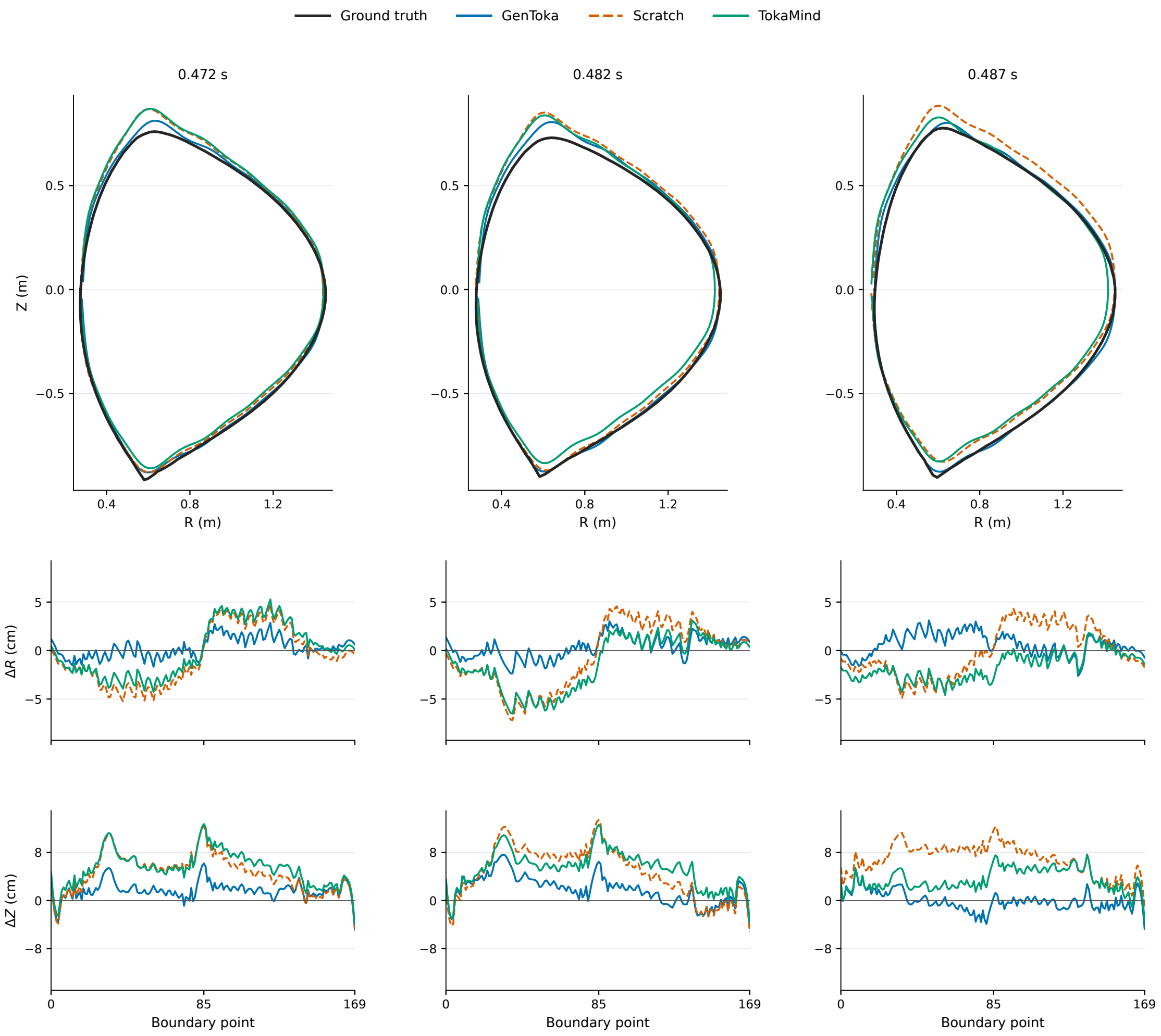}
    \caption{During magnetic-dynamics forecasting, GenToka follows the separatrix in $(R,Z)$ more tightly than TokaMind on this discharge, echoing the largest single-task NRMSE gain in the family.}
    \label{fig:qual-magnetic}
\end{figure}

\subsubsection{Profile and Diagnostic Dynamics}
\label{sec:profile-results}
The profile and diagnostic tasks extend the comparison to a different set of temporal and spatial outputs. GenToka lowers group-mean NRMSE from 0.2807 to 0.2660 relative to TokaMind (5.2\%). Task 3-2 changes from 0.2529 to 0.2522, while Task 3-3 changes from 0.2415 to 0.2147, locating the larger gain within the family.
Relative to Scratch, pretraining lowers the group mean from 0.2888 to 0.2660 (7.9\%); the clearest separation is on Task~3-2, where NRMSE falls from 0.2943 to 0.2522 (14.3\%), suggesting that reused cross-signal representations matter on profile-heavy forecasting targets.
\begin{figure}[!htpb]
    \centering
    \includegraphics[page=14,width=0.9\linewidth,keepaspectratio]{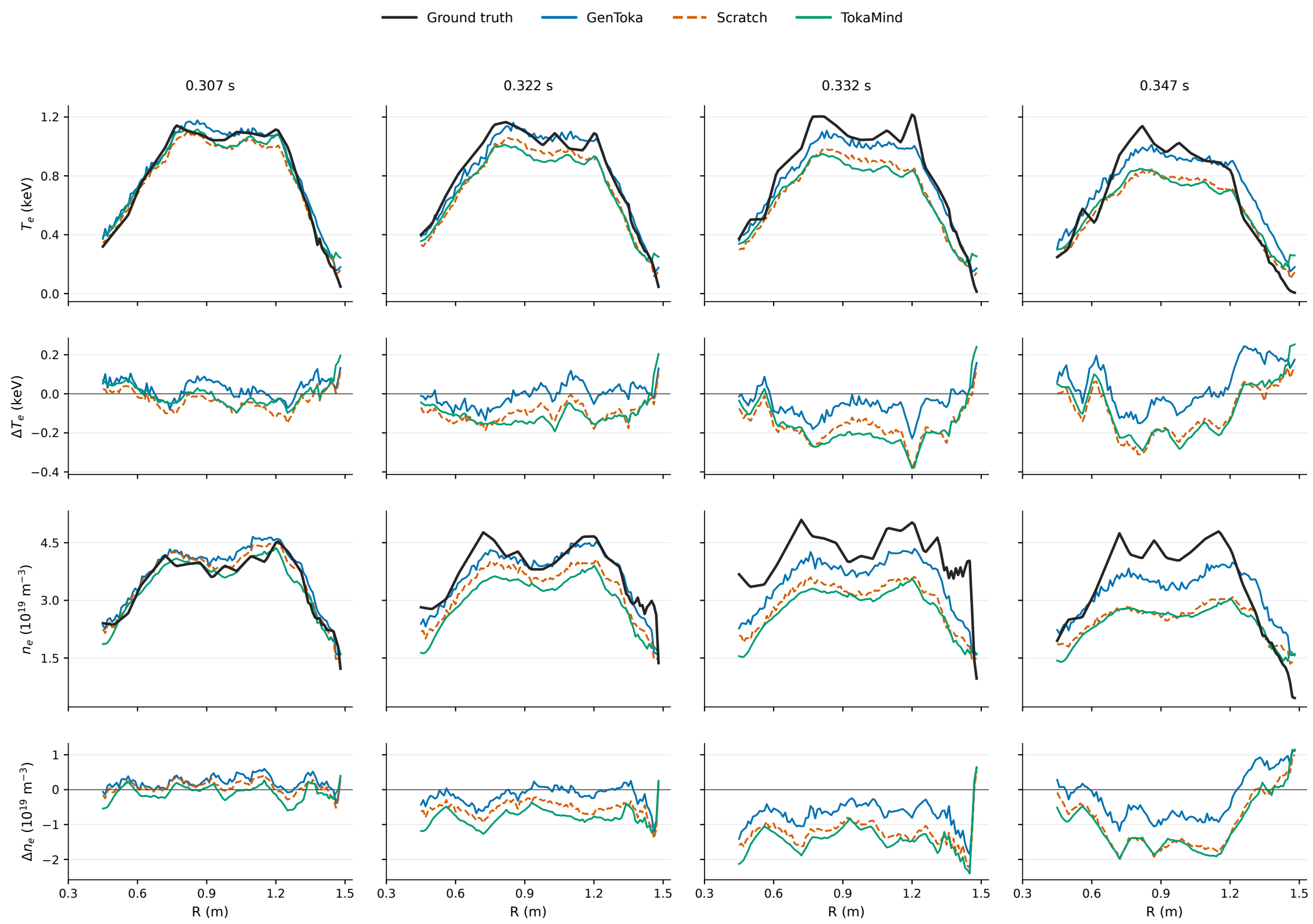}
    \caption{When predicting kinetic profiles ahead in time, GenToka produces smoother, better-aligned $T_e$ and $n_e$ than TokaMind, consistent with the stronger reductions on the most profile-heavy targets.}
    \label{fig:qual-profile}
\end{figure}

\subsubsection{Long-Horizon MHD Dynamics}
\label{sec:long-horizon-results}
For the extended-horizon schemas, GenToka and TokaMind have the same group-mean NRMSE at the displayed precision (0.3179). GenToka has lower error on Tasks 4-3 and 4-4, while TokaMind has lower error on Tasks 4-1, 4-2, and 4-5.
Scratch averages 0.3325 NRMSE on this group; pretraining lowers the group mean to 0.3179 (4.4\%) even though Scratch is slightly better on Task~4-4, so the long-horizon tie with TokaMind still reflects a gain over random initialization.
\begin{figure}[!htpb]
    \centering
    \includegraphics[page=18,width=0.9\linewidth,keepaspectratio]{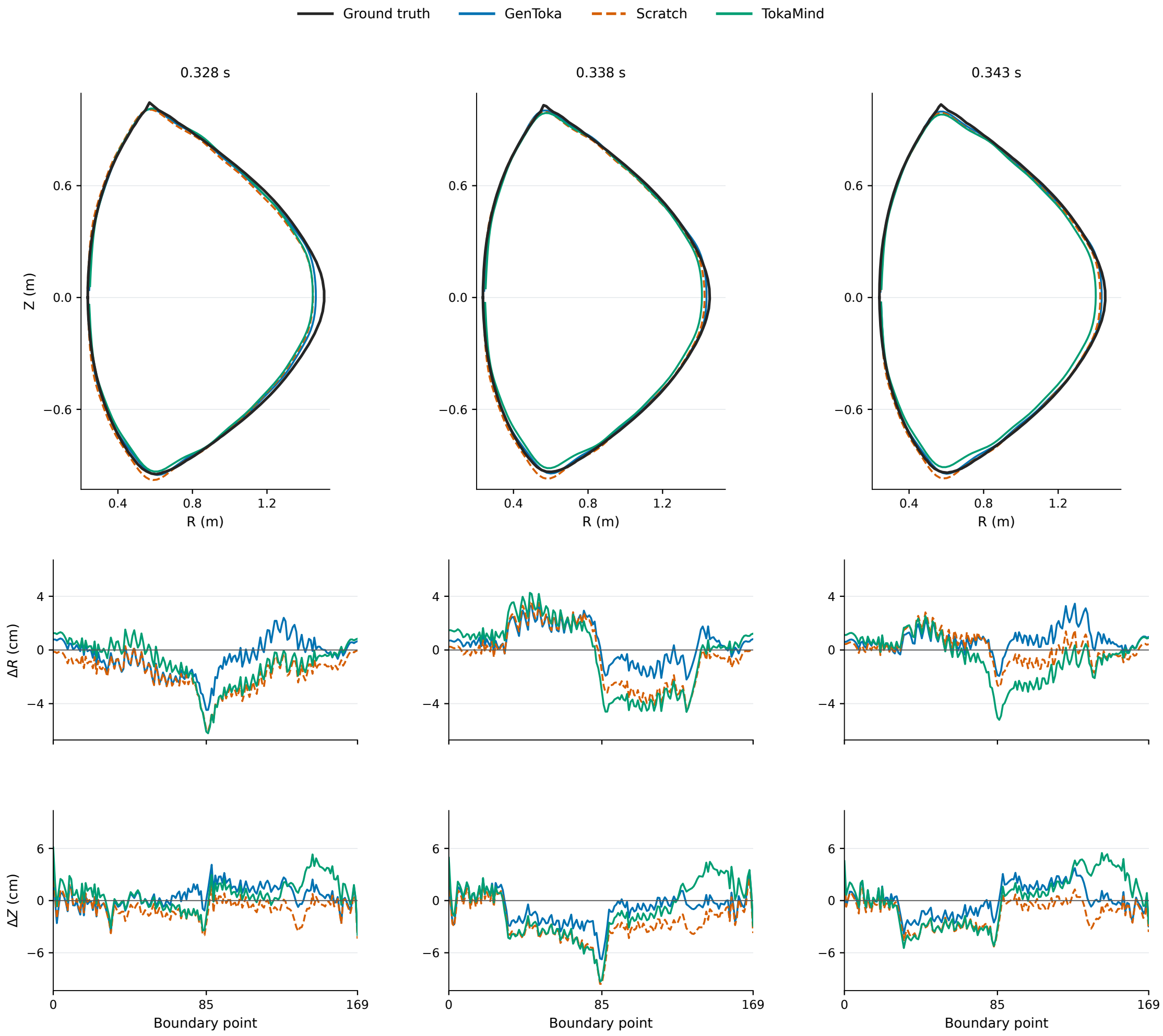}
    \caption{At a 100\,ms horizon the two models diverge on magnetic-axis and separatrix geometry on the same shot, explaining why long-horizon averages can tie even when individual targets favor different models.}
    \label{fig:qual-longhorizon}
\end{figure}

\subsubsection{Actuator Action Inversion}
\label{sec:inverse-results}

Actuator action inversion evaluates the recovery of recorded control-signal time series from historical measurements and a future plasma response observed in the same discharge. Across Tasks 5-1--5-3, GenToka lowers NRMSE relative to TokaMind by 45.9\%, 4.4\%, and 19.1\%, respectively, giving a 17.7\% reduction in the group mean. Table~\ref{tab:inverse-components} reports the corresponding native-space errors for all sixteen target components. These TokaMind--GenToka results use the complete pipeline settings specified in the comparison protocol.
Pretraining lowers the group mean from 0.2130 to 0.1946 (8.6\%) relative to Scratch, with GenToka ahead on all three inverse tasks, indicating that foundation pretraining helps relate observed responses to control trajectories bidirectionally rather than relying on task-only training alone.
\begin{figure}[!htpb]
    \centering
    \includegraphics[page=22,width=0.9\linewidth,keepaspectratio]{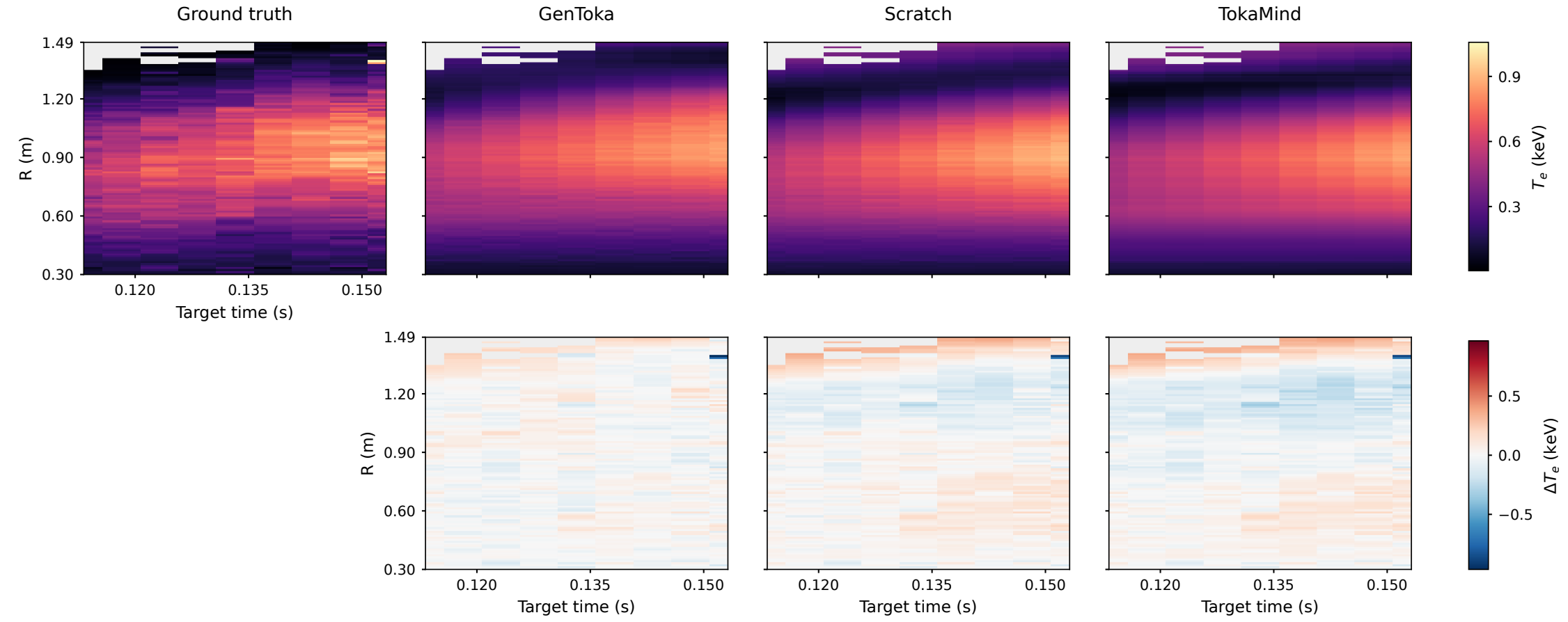}
    \caption{Given an observed plasma response in the same discharge, GenToka reconstructs the recorded PF-coil voltages more faithfully than TokaMind, in line with the large gains on magnetic actuation inversion.}
    \label{fig:qual-actuator}
\end{figure}

\begin{figure}[!htpb]
    \centering
    \includegraphics[width=0.9\linewidth]{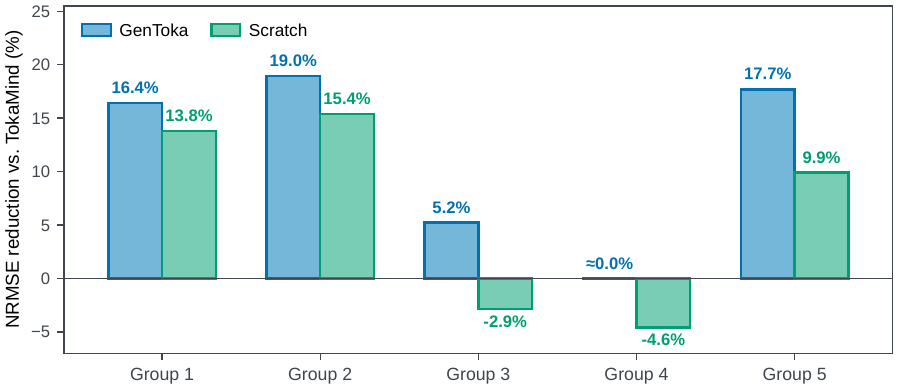}
    \caption{Relative test-NRMSE improvement of GenToka over TokaMind by task family (positive is better): GenToka leads on equilibrium, magnetic, profile, and actuator-inversion families, ties on long-horizon MHD, and Scratch without pretraining is consistently weaker.}
    \label{fig:group-level-nrmse-improvement}
\end{figure}
\noindent Figure~\ref{fig:group-level-nrmse-improvement} aggregates the same pattern across Groups~1--5: GenToka lowers group-mean NRMSE relative to TokaMind in Groups~1--3 and Group~5, with matched displayed means in Group~4.

\subsubsection{Disruption Prediction}
\label{sec:disruption-results}

Group 6 evaluates disruption prediction using nested sets of historical inputs: nine signals in Task 6-1, five in Task 6-2, and plasma current alone in Task 6-3 (Table~\ref{tab:expanded-task-schemas}). AUROC measures window-level risk ranking, and F1 measures classification at the common validation-selected threshold $\tau^\star=0.4$.

GenToka achieves higher F1 than TokaMind on all three tasks: 0.6297, 0.6206, and 0.4674, compared with 0.6029, 0.5923, and 0.3891. The corresponding absolute gains are 0.0268, 0.0283, and 0.0783, increasing macro-average F1 from 0.5281 to 0.5726 (8.4\%). The largest gain occurs on Task 6-3, which uses plasma current alone; GenToka also achieves higher AUROC on this task (0.8824 versus 0.8683). TokaMind has higher AUROC on Tasks 6-1 and 6-2, with macro-averages of 0.9069 for TokaMind and 0.9049 for GenToka across all three tasks. The consistent F1 improvement therefore identifies an advantage at the shared classification threshold, including the single-signal setting.
Scratch macro-average F1 is 0.5588 under the same threshold; pretraining raises it to 0.5726 (2.5\%), with the largest relative gain again on Task~6-3, where F1 increases from 0.4427 to 0.4674 (5.6\%).
AUROC is already high for every model (roughly 0.87--0.93), so GenToka can be slightly below TokaMind or Scratch on Tasks~6-1 and 6-2 without a large practical gap in ranking.
\begin{figure}[!htpb]
    \centering
    \includegraphics[page=27,width=0.9\linewidth,keepaspectratio]{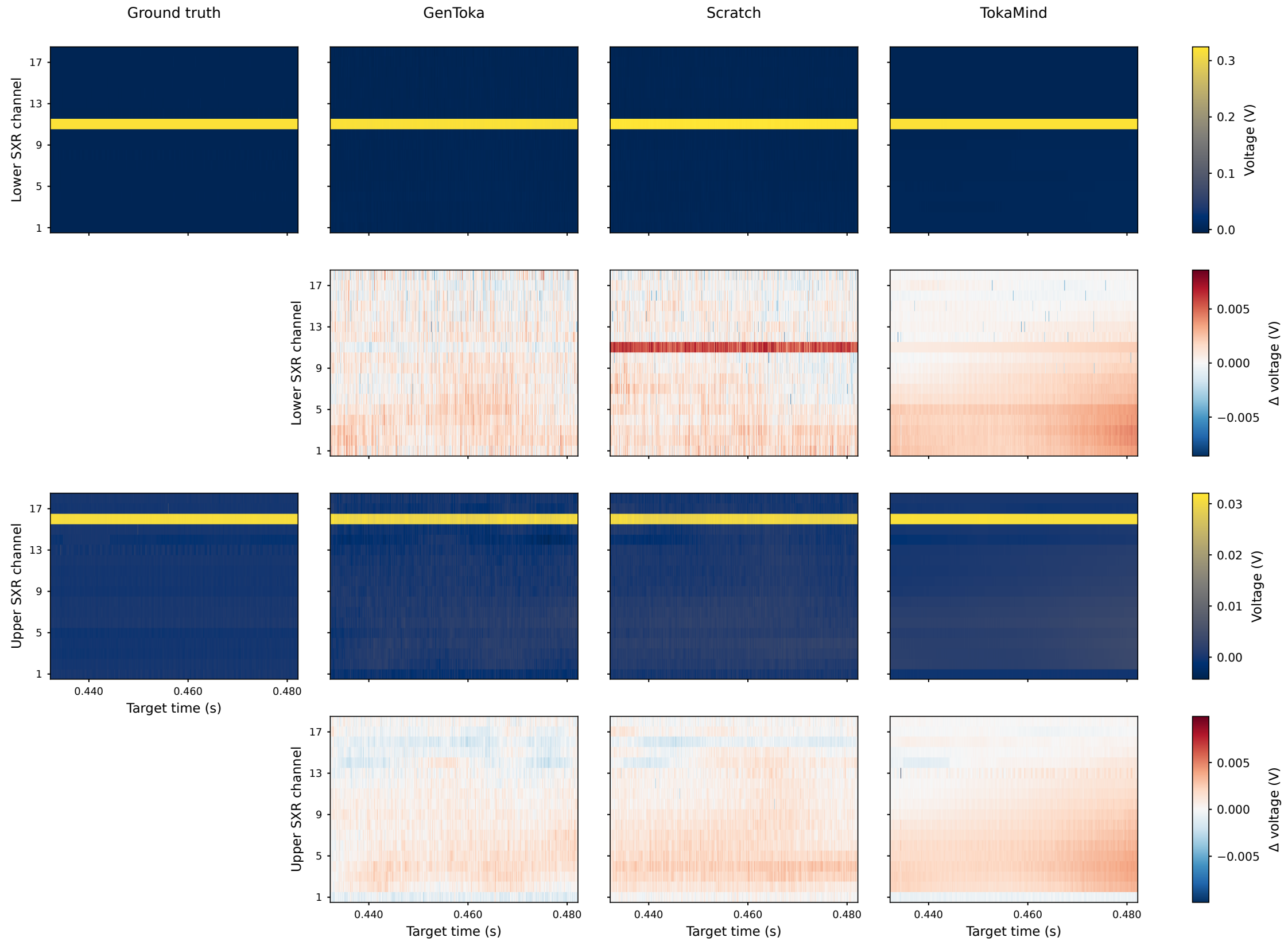}
    \caption{Near a disruption-warning window with multivariate history, GenToka assigns clearer separation at the shared risk threshold than TokaMind, matching the improved F1 despite similar AUROC.}
    \label{fig:qual-disruption}
\end{figure}
\section{Discussion}
\label{sec:discussion}

The MAST evaluation suggests that a schema-flexible generative interface can cover the same breadth as deterministic multimodal predictors while improving accuracy on several task families. Relative to TokaMind, GenToka lowers macro-average error on plasma-response and actuator-inversion groups and raises disruption F1 at a shared operating threshold, indicating that joint target-space denoising and foundation pretraining complement the TokaMind-style token organization rather than replacing the need for careful task specification. Paired comparisons with Scratch further show that initialization matters: the same architecture and downstream recipe benefit from pretraining on co-temporal signals before schema-specific adaptation, especially where tasks require relating many diagnostics to heterogeneous outputs.

Three limitations point to directions for extending this line of work. \textbf{Non-autoregressive multi-step prediction.} Each task generates a complete output window in one conditional pass rather than advancing time autoregressively with updated context. This matches the fixed-horizon offline schemas used here and in TokaMark, but it restricts direct deployment in settings where the model must be applied repeatedly along a discharge, for example, rolling forecasting, iterative reconstruction, or closed-loop control.

\textbf{Actuator inversion versus unified control modeling.} Group~5 focuses on recovering recorded control-signal trajectories for a given actuator pathway from an observed plasma response. A next step is to cast control objectives as additional schemas within the same generative tokamak foundation interface, so that forward modeling, inversion, disruption risk, and control sit in one conditional framework, as unified multi-task generative models have demonstrated in other physical domains \citep{feng2025fluidzero}. Concretely, control tasks could be realized by guiding generation on explicit control targets, such as desired plasma states, tracking goals, or stability margins, while reusing the shared denoising backbone and signal-specific decoders.

\textbf{Single-equipment training.} Foundation pretraining and downstream adaptation in this study use MAST data only; unification is therefore across tasks and signal roles on one equipment rather than across tokamak parameters, coil configurations, and diagnostic inventories. Incorporating additional equipment with harmonized multimodal records would test cross-equipment transfer and clarify whether schema-flexible pretraining can support equipment-agnostic tokamak foundation models under realistic distribution shift.

\section{Conclusion}
\label{sec:conclusion}

This work presents an extended tokamak task benchmark and GenToka, a generative foundation model that adapts to schema-defined inputs and outputs, with the goal of unifying plasma-response forecasting, actuator action inversion, and disruption risk under one multimodal interface on experimental measurements. We extend TokaMark with other two groups of tasks in Groups~5--6, yielding 20 tasks in six groups with explicit schemas, shared evaluation protocols, and complete LSTM and TokaMark-CNN reference results. GenToka builds on the TokaMind token organization and replaces deterministic prediction with joint conditional diffusion over heterogeneous targets, using signal-specific codecs and a shared Transformer denoiser, with foundation pretraining followed by task-specific adaptation. On the MAST data, GenToka reduces macro-average NRMSE relative to TokaMind by 4.7\% on Groups~1--4 and by 17.7\% on Group~5, improves macro-average window-level disruption F1 by 8.4\% at a common validation-selected threshold, and outperforms the same architecture initialized from scratch, indicating benefits from both the generative formulation and foundation pretraining. Together, the benchmark and model provide a reusable and generative pathway for integrated tokamak modeling.

\subsubsection*{Acknowledgments}

This  work  is  supported  by  the  Zhongguancun  Academy  (Grant  Nos.  C20250518, XTS0080). We would also like to thank the development groups of TokaMark and TokaMind for open-sourcing their datasets and code, which provided the foundation for this work.

\clearpage
\bibliographystyle{iclr2026_conference}
\bibliography{iclr2026_conference}

\clearpage
\appendix
\section{Additional Experimental Protocols}
\label{sec:appendix-experimental-details}

\subsection{Schemas of the Expanded Tasks}
\label{sec:appendix-task-schemas}

Table~\ref{tab:expanded-task-schemas} records the observed context, paired future information, requested targets, and temporal support of each extension. The three risk configurations have nested historical inputs and identical test shot identities, cut times, window indices, and Soft-TTD targets.

\begin{table*}[!ht]
\centering
\caption{Schemas of the six extension tasks. Windows are relative to the
cut time $t_c$ and are left-closed, right-open. Task~5 conditions on a paired
observed future plasma response to reconstruct the corresponding actuator time series.
Task~6 uses causal history to predict a continuous soft time-to-disruption
target.}
\label{tab:expanded-task-schemas}
\scriptsize
\setlength{\tabcolsep}{3pt}
\renewcommand{\arraystretch}{1.16}
\begin{tabular}{>{\raggedright\arraybackslash}p{0.055\textwidth}>{\raggedright\arraybackslash}p{0.285\textwidth}>{\raggedright\arraybackslash}p{0.205\textwidth}>{\raggedright\arraybackslash}p{0.17\textwidth}>{\raggedright\arraybackslash}p{0.19\textwidth}}
\toprule
Task & Observed history & Observed future response & Prediction target & Window and held-out cohort \\
\midrule
5-1 & Magnetic diagnostics; PF-coil, solenoid and plasma currents; past PF voltages; optional NBI power.
& P2/P4/P5 coil currents, solenoid current and $I_p$; optional LCFS $(R,Z)$ contour.
& Four PF voltage traces: P1, P2, P4, P5.
& History: 5\,ms; response/target: 25\,ms; 1\,ms cut grid; 1,103 test shots. \\
5-2 & Magnetic diagnostics; PF-coil, solenoid and plasma currents; line-integrated density; Thomson $T_e,n_e$; past current/density requests, NBI power, cumulative gas and gas-valve voltages.
& Thomson electron-density profile.
& Six gas-valve voltage traces: G1, G2, G3, G4, G11, G12.
& History: 150\,ms (Thomson: 5\,ms); response/target: 50\,ms; 1\,ms cut grid; 681 test shots. \\
5-3 & Thomson $T_e,n_e$; D-alpha and upper/lower soft-X-ray camera voltages; past current/density requests, NBI power, cumulative gas and gas-valve voltages.
& Three-channel D-alpha voltage time series.
& Six gas-valve voltage traces: G1, G2, G3, G4, G11, G12.
& History: 5\,ms; response/target: 50\,ms; 1\,ms cut grid; 803 test shots. \\
6-1 & Nine signals: $I_p$, line-integrated density, radiated power, D-alpha voltage, lower soft-X-ray camera voltage, $q_{95}$, internal inductance $l_i$, diamagnetic energy and minor radius.
& ---
& Soft time-to-disruption risk.
& History: 100\,ms; 10\,ms cut grid; 933 test shots / 39,967 windows. \\
6-2 & Five signals: $I_p$, line-integrated density, D-alpha voltage, lower soft-X-ray camera voltage and minor radius.
& ---
& Soft time-to-disruption risk.
& Same history, grid, shots, window coordinates and targets as 6-1. \\
6-3 & Plasma current $I_p$.
& ---
& Soft time-to-disruption risk.
& Same history, grid, shots, window coordinates and targets as 6-1. \\
\bottomrule
\end{tabular}
\end{table*}

\FloatBarrier

\subsection{Backbone and Paired Initialization Protocols}
\label{sec:appendix-model-protocols}

The DCT-based actuator action inversion and disruption-risk models use twelve Transformer blocks with hidden width 384, six attention heads, and feed-forward width 1536. Full attention operates over the context and target tokens, with timestep-conditioned AdaLN. Diffusion operates on target codec representations, while individual conditioning signals use their schema-defined DCT or identity codecs.

For each Group~1--4 task, pretrained GenToka and Scratch use the same architecture and downstream recipe, differing in initialization. The Group~5 pairs also share the frozen task schema, preprocessing, codec supports, optimizer settings, and training windows. Both Group~5 models use AdamW, batch size 2048, and 100 epochs: 25 epochs of interface alignment with the backbone frozen, followed by 75 epochs with the context tokenizer frozen. The second-stage backbone learning rate is $10^{-3}$, with a six-epoch backbone warmup. The reported Group~5 evaluations use the final epoch-100 checkpoint. Diffusion uses 1000 cosine-schedule training levels, velocity prediction, signal-scalar target normalization, and two deterministic implicit sampling steps ($\eta=0$). The frozen seed is 54.

For disruption prediction, GenToka and Scratch use AdamW, batch size 2048, and the same two-stage 25+75-epoch structure, with second-stage backbone learning rate $7.5\times10^{-4}$ and a six-epoch backbone warmup. Both use balanced Soft-TTD training strata and the same selected training windows. Their checkpoints are selected by sampled native-space validation NRMSE on a fixed, shot-stratified set of 3000 validation windows. The risk models likewise use two implicit sampling steps with $\eta=0$. The TokaMind--GenToka comparisons evaluate their complete training and prediction pipelines; the paired GenToka--Scratch comparisons hold the downstream recipe fixed.

\subsection{Native-Space Actuator Action Inversion Component Results}
\label{sec:appendix-inverse-components}

The Group~5 output roles cover four PF-coil voltage channels in Task 5-1 and six gas-valve voltage channels in each of Tasks 5-2 and 5-3. Table~\ref{tab:inverse-components} reports each component on the same frozen accepted test cohort and native-space normalization used by its matched TokaMind--GenToka comparison. GenToka has lower NRMSE on all sixteen components.

\begin{table}[!ht]
\centering
\small
\caption{Decoded native-space component NRMSE on the frozen Group~5 test cohorts. Each component uses the shared signal-role standard deviation and equal weighting across accepted test shots. Relative reduction is measured against TokaMind. Component averages differ from task NRMSE because task NRMSE pools output components within each shot before averaging across shots.}
\label{tab:inverse-components}
\begin{tabular}{llrrr}
\toprule
Task & Target component & TokaMind & GenToka & Reduction (\%) \\
\midrule
5-1 & XDC\_PF\_F\_P1 & 0.1545 & 0.0706 & 54.3 \\
5-1 & XDC\_PF\_F\_P2 & 0.0710 & 0.0360 & 49.3 \\
5-1 & XDC\_PF\_F\_P4 & 0.1437 & 0.0825 & 42.6 \\
5-1 & XDC\_PF\_F\_P5 & 0.1507 & 0.0881 & 41.5 \\
5-2 & G1 & 0.2930 & 0.2842 & 3.0 \\
5-2 & G2 & 0.2874 & 0.2761 & 3.9 \\
5-2 & G3 & 0.2921 & 0.2808 & 3.8 \\
5-2 & G4 & 0.2982 & 0.2850 & 4.4 \\
5-2 & G11 & 0.3110 & 0.2950 & 5.2 \\
5-2 & G12 & 0.3130 & 0.2960 & 5.4 \\
5-3 & G1 & 0.2327 & 0.1910 & 17.9 \\
5-3 & G2 & 0.2304 & 0.1863 & 19.1 \\
5-3 & G3 & 0.2312 & 0.1888 & 18.3 \\
5-3 & G4 & 0.2344 & 0.1889 & 19.4 \\
5-3 & G11 & 0.2444 & 0.1918 & 21.5 \\
5-3 & G12 & 0.2450 & 0.1944 & 20.6 \\
\bottomrule
\end{tabular}
\end{table}

\FloatBarrier

\subsection{Benchmark Reference Architectures and Training}
\label{sec:appendix-benchmark-baselines}

The benchmark reference suite contains forty trained models: one LSTM and one TokaMark-CNN for each of the twenty tasks. Every model is initialized randomly and trained for 50 epochs. Table~\ref{tab:baseline-training} summarizes the optimization settings; task schemas, shot partitions, conditioning-time restrictions, and native-space evaluation rules are shared across the reference implementations. The reported CNN results are reruns under these benchmark schemas and settings, including the six extension tasks.

\subsubsection{LSTM}
\label{sec:appendix-recurrent-baseline}

The LSTM uses the same retained DCT coefficients as GenToka: truncated inverse DCT reconstructs each allowed conditioning chunk, which is encoded along physical time by a two-layer LSTM; a second two-layer LSTM processes the chunk sequence independently for each role--signal pair before fusion and coefficient-space regression. Signals assigned identity codecs retain their native values. Within-chunk sequences are reduced to at most 32 temporal samples. The within-chunk and across-chunk hidden widths are 16 and 32, respectively, followed by a 128-dimensional fusion network. Predictions are decoded into native signal space for evaluation; the scalar Soft-TTD target uses its identity codec.

\subsubsection{TokaMark-CNN}
\label{sec:appendix-cnn-baseline}

The convolutional reference retains the multi-branch encoder--fusion--decoder structure of the TokaMark baseline \citep{rousseau2026tokamark}. Each signal branch operates on standardized native arrays, using one-, two-, or three-dimensional convolutions according to the native temporal and spatial axes. Each encoder has three convolutional stages with kernel size 3, stride 3, padding 1, ReLU, max pooling, and batch normalization. The base width is 16, increasing to 32 and 64 channels; a linear projection produces a 32-dimensional branch representation. A two-layer MLP fuses historical branch representations. Schema-permitted future conditioning signals are encoded by separate branches and incorporated through a second fusion MLP.

Each requested output has a three-stage transposed-convolution decoder with kernel size 3 and stride 3, with per-layer cropping to the required native shape. The fused 32-dimensional representation is repeated across output chunks, retaining the TokaMark decoder structure. Task schemas route the observed response into the conditioning branches for actuator action inversion and restrict disruption inputs to pre-anchor history. A sigmoid maps the scalar disruption output to the Soft-TTD range. Native input and output arrays bypass the DCT representation; missing conditioning slots are zero-filled after standardization, and supervised losses respect the target masks.

\begin{table}[!ht]
\centering
\small
\caption{Training settings for the two benchmark reference implementations. Both use seed 54, eight GPUs, a 2\% linear learning-rate warmup, and 3000 fixed shot-stratified validation windows for checkpoint selection. The reported result for each task uses its best validation checkpoint.}
\label{tab:baseline-training}
\setlength{\tabcolsep}{5pt}
\begin{tabular}{lll}
\toprule
Setting & LSTM & TokaMark-CNN \\
\midrule
Training epochs & 50 & 50 \\
Optimizer & AdamW & Adam \\
Learning rate & $6\times10^{-4}$ & $3\times10^{-4}$ \\
Weight decay & $10^{-2}$ & 0 \\
Global batch size & 1024 & 2048 \\
Supervision / selection & Masked coefficient MSE & Masked native-space MSE \\
\bottomrule
\end{tabular}
\end{table}

All seventeen continuous-output tasks use the same accepted test shots, target entries, per-target window counts, training-set normalization, and shot-weighted native NRMSE aggregation as the frozen benchmark evaluator. The three disruption tasks use the nested nine-, five-, and one-signal histories, balanced Soft-TTD training strata, and the common held-out window grid. LSTM and TokaMark-CNN both use the frozen F1 threshold $\tau^\star=0.4$ and do not participate in threshold selection. The reference results describe the two complete training recipes; their input representations, losses, and optimization settings are specified separately.

\FloatBarrier

\subsection{Validation of the Common Disruption Threshold}
\label{sec:appendix-threshold-selection}

Equation~\ref{eq:disruption-threshold} selects a common threshold using only the six primary validation sets. The threshold is frozen before test evaluation; Scratch, LSTM, and TokaMark-CNN models do not participate in its selection.

Table~\ref{tab:threshold-validation} reports the validation statistic used by this rule. Each entry is the unweighted mean F1 over the six primary model--task pairs.

\begin{table}[!ht]
    \caption{Validation macro-average F1 for the common disruption-risk threshold. The average covers TokaMind and GenToka on Tasks 6-1--6-3; no test labels are used. Bold entries identify the selected threshold and its validation score.}
    \label{tab:threshold-validation}
    \centering
    \small
    \setlength{\tabcolsep}{4pt}
    \begin{tabular}{@{}lrrrrrrrrr@{}}
        \toprule
        $\tau$ & 0.1 & 0.2 & 0.3 & \textbf{0.4} & 0.5 & 0.6 & 0.7 & 0.8 & 0.9 \\
        \midrule
        Mean F1 & 0.4105 & 0.4884 & 0.5391 & \textbf{0.5570} & 0.5226 & 0.4675 & 0.3932 & 0.2983 & 0.1724 \\
        \bottomrule
    \end{tabular}
\end{table}

The mean F1 increases through $\tau=0.4$ and decreases thereafter, making $\tau^\star=0.4$ the unique maximizer on the prespecified grid. At the selected operating point, the six constituent validation F1 values are $0.6165$, $0.5993$, and $0.4056$ for TokaMind on Tasks 6-1--6-3, and $0.6246$, $0.6177$, and $0.4785$ for GenToka on the same tasks, respectively. Their unweighted mean is $0.5570$, which is the statistic used to freeze the test-time threshold.

\end{document}